\pdfoutput=1

\documentclass[11pt]{article}

\usepackage[preprint]{acl}

\makeatletter
\AtBeginDocument{%
  \ifdefined\@LN@orig@makecol
    \IfFormatAtLeastTF{2025-06-01}{%
      \RemoveFromHook{build/column/before}[lineno]%
      \def\@LN@makecol{%
        \@LN@orig@makecol
        \setbox\@outputbox \vbox{%
          \boxmaxdepth \@maxdepth
          \protected@write\@auxout{}{%
            \string\def\string\@LN@column{\if@firstcolumn1\else2\fi}%
          }%
          \box\@outputbox
        }%
      }%
      \let\@makecol\@LN@makecol
    }{}%
  \fi
}
\makeatother

\usepackage{times}
\usepackage{latexsym}
\usepackage{amssymb}
\usepackage{amsmath}
\usepackage{pifont}
\usepackage{float}

\newcommand{\cmark}{\ding{51}}

\newcommand{\xmark}{\ding{55}}
\usepackage{enumitem}

\usepackage[T1]{fontenc}

\usepackage[utf8]{inputenc}
  
\usepackage{microtype}

\usepackage{inconsolata}

\usepackage{graphicx}
\usepackage{fontawesome5}
\usepackage{simpleicons}
\newcommand{\hficon}{\raisebox{-0.1em}{\simpleicon{huggingface}}}
\newcommand{\mosaiclinkitem}[3]{%
  \href{#1}{%
    \textcolor{black}{\raisebox{-0.06ex}{\makebox[1.15em][c]{#2}}}%
    \hspace{0.28em}%
    \textcolor{darkblue}{\textbf{#3}}%
  }%
}
\usepackage{multirow}
\usepackage{booktabs}
\usepackage{xcolor}
\usepackage{placeins}
\usepackage{capt-of}
\usepackage{tcolorbox}
\tcbuselibrary{skins,breakable}

\tcbset{aibox/.style={
    unbreakable,
    fontlower=\scriptsize,
}}

\newtcolorbox{AIbox}[2][]{aibox,title={#2},#1}

\title{MOSAIC: An Omni-Modal Benchmark for Long-Term Memory and Cross-Modal Reasoning in Conversational Agents}

\author{%
  \textbf{Hong Ting Tsang}$^{1}$, \textbf{Boyi Zhang}$^{1}$, \textbf{Haoyu Huang}$^{1}$, \textbf{Jiaxin Bai}$^{1}$, \textbf{Yufei Li}$^{1}$, \\
  \textbf{Yisen Gao}$^{1}$, \textbf{Zhongwei Xie}$^{1}$, \textbf{Hongyu Luo}$^{1}$, \textbf{Jeff Z.\ Pan}$^{2}$, \textbf{Yangqiu Song}$^{1}$ \\[0.3em]
  {\normalfont $^{1}$The Hong Kong University of Science and Technology,\quad
  $^{2}$University of Edinburgh} \\[0.2em]
  {\normalfont\small \texttt{\{httsangaj, bzhangdd, hhuangcp\}@connect.ust.hk},\ \texttt{baijiaxin@hkbu.edu.hk}} \\
  {\normalfont\small \texttt{\{ylivm, ygaodi, zxiebk, hluoay\}@connect.ust.hk},\ \texttt{j.z.pan@ed.ac.uk},\ \texttt{yqsong@cse.ust.hk}} \\[0.25em]
  {\normalfont\small
    \mosaiclinkitem{https://github.com/HKUST-KnowComp/MOSAIC}{\faGithub}{Code}%
    \hspace{0.75em}{\textcolor{black!35}{\raisebox{0.12ex}{\textperiodcentered}}}\hspace{0.75em}%
    \mosaiclinkitem{https://huggingface.co/datasets/gzone0111/MOSAIC}{\hficon}{Dataset}%
    \hspace{0.75em}{\textcolor{black!35}{\raisebox{0.12ex}{\textperiodcentered}}}\hspace{0.75em}%
    \mosaiclinkitem{https://hkust-knowcomp.github.io/MOSAIC/}{\faGlobe}{Website}}%
}

\begin{document}
\maketitle
\begin{abstract}
Omni-modal assistants must remember not only what users say, but also what they show and convey. Existing long-term memory benchmarks are mostly text-and-image or report a single end-to-end score, so they cannot locate whether evidence is lost when it is \emph{represented}, \emph{retrieved}, \emph{delivered}, or \emph{read}. We introduce \textbf{MOSAIC}, a benchmark of 923 questions over persona-grounded histories of 11--24 sessions spanning text, image, video, and audio. MOSAIC constructs \emph{targeted} spatial, temporal, and prosodic signals rather than decorative media---with the strongest causal validation for spatial information and selector-level validation for annotated prosody---and pairs them with a matched diagnostic protocol that freezes the question, reader, prompt, and budget while intervening on representation, retrieval, and delivery. Matched decomposition shows that representation loss is modality-dependent, residual reading dominates even under gold evidence, and abstention remains weak on these controls: strong answerable scores coexist with failures to refuse. MOSAIC thus turns omni-modal memory evaluation from a leaderboard into a diagnosis of \textsc{write}, \textsc{retrieve}, \textsc{deliver}, and \textsc{read} failures.
\end{abstract}

\section{Introduction}
\label{sec:intro}

\begin{figure}[t]
    \centering
    \includegraphics[width=\linewidth]{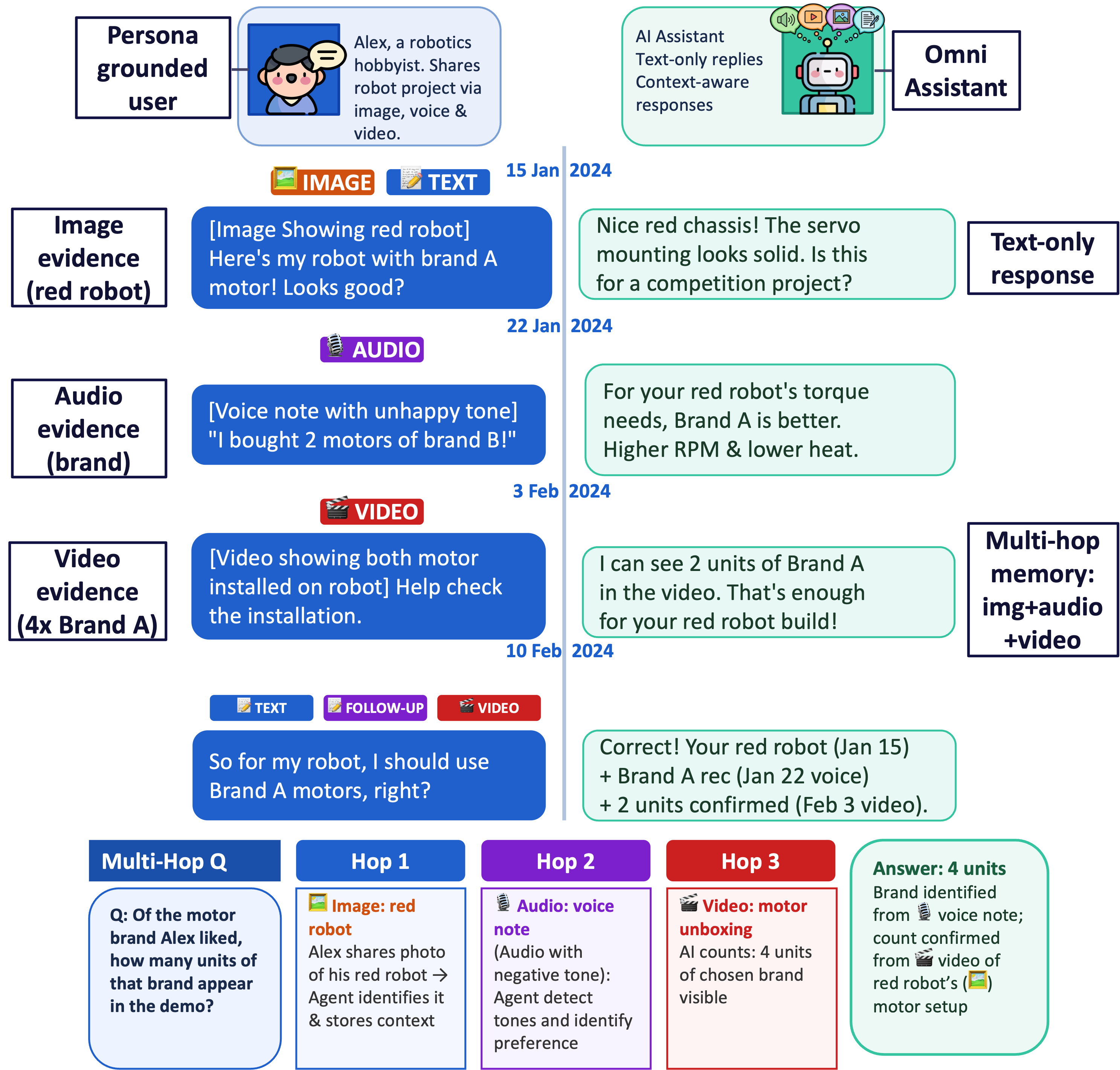}
    \caption{Example MOSAIC haystack: evidence spans four modalities and months of sessions. Answering the final query requires chaining a January text specification with later audio and video evidence.}
    \label{fig:example-haystack}
    \vspace{-0.2in}
\end{figure}

Omni-modal assistants such as GPT-4o, Gemini, Qwen-Omni, and MiMo~\citep{openai_gpt-4o_2024, team_gemini_2025, qwen_team_qwen35-omni_2026, noauthor_xiaomimimomimo-v25_2026} are moving from single-turn demos to persistent, multi-session deployments. A user may share a product photo in January, leave a dietary voice note in March, and later expect both to inform a recommendation. Answering requires not only retrieving temporally scattered evidence, but \emph{chaining it across modalities} (Figure~\ref{fig:example-haystack}). Prior conversational memory benchmarks omit video/audio, treat images as optional, or stop at text--image co-reference; none combine four-modality multi-session dialogue with a matched stage-diagnostic protocol (Table~\ref{tab:benchmark-comparison}).

\begin{figure*}[t]
  \centering
  \includegraphics[width=\linewidth]{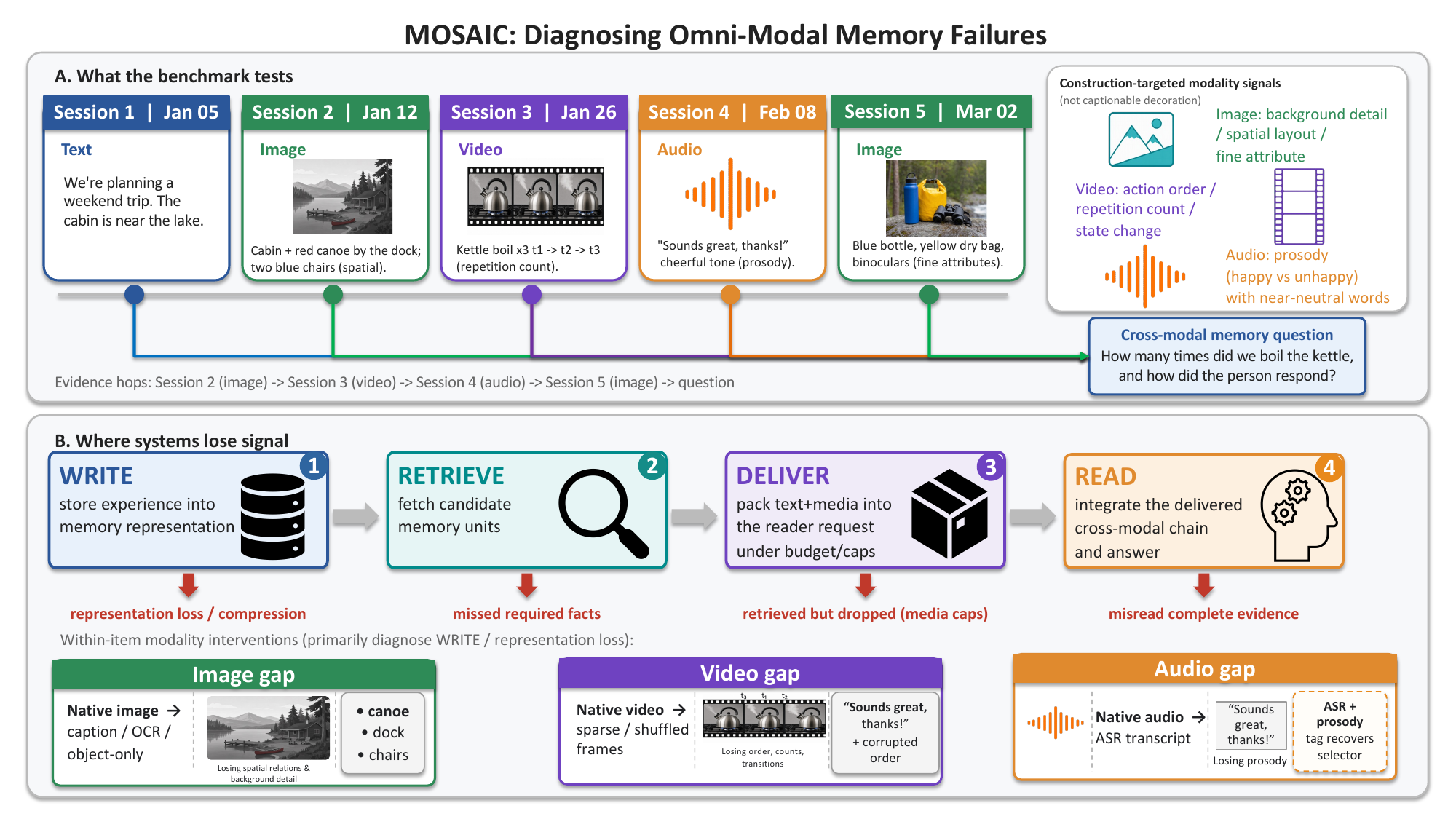}
  \caption{Overview of MOSAIC. \textbf{Top:} what the benchmark tests---evidence
  scattered across months of text, image, video, and audio sessions must be
  chained across modalities. Construction targets native signals rather than
  decorative media: spatial layout and background detail in images; action
  order, repetition count, and state change in video; vocal affect in audio
  (causal validation is strongest for spatial detail; \S\ref{sec:results:modality}).
  \textbf{Bottom:} where systems lose that signal---matched interventions on
  \textsc{write} (representation compression), \textsc{retrieve} (missed
  facts), \textsc{deliver} (retrieved but dropped under media caps), and
  \textsc{read} (misread complete evidence).}
  \label{fig:overview}
\end{figure*}

\begin{table*}[t]
\centering
\small
\setlength{\tabcolsep}{4pt}
\begin{tabular}{@{}lcccccccrr@{}}
\toprule
& & \multicolumn{4}{c}{\textbf{Modalities}} & \multicolumn{2}{c}{\textbf{Reasoning}} & & \\
\cmidrule(lr){3-6}\cmidrule(lr){7-8}
\textbf{Benchmark} & \textbf{Multi-Sess.} & T & I & V & A & \textbf{Multi-Hop} & \textbf{Cross-Mod.} & \textbf{\#Items} & \textbf{Avg.\ Sess.}\\
\midrule
MemoryBank~\citep{zhong_memorybank_2023}        & \cmark & \cmark & \xmark & \xmark & \xmark & \xmark & \xmark &  194 & 3.8 \\
DialogBench~\citep{ou_dialogbench_2024}         & \cmark & \cmark & \xmark & \xmark & \xmark & \xmark & \xmark & 9{,}789 & 7.5 \\
LongMemEval~\citep{wu_longmemeval_2025}         & \cmark & \cmark & \xmark & \xmark & \xmark & \cmark & \xmark &  500 & 5.2 \\
MemAgentBench~\citep{yu_memagent_2025}  & \cmark & \cmark & \xmark & \xmark & \xmark & \cmark & \xmark & 1{,}200 & 9.6 \\
\midrule
LoCoMo~\citep{maharana_evaluating_2024}         & \cmark & \cmark & \cmark & \xmark & \xmark & \xmark & \xmark & 1{,}986 & 27.2 \\
MMDU~\citep{liu_mmdu_2024}                      & \xmark & \cmark & \cmark & \xmark & \xmark & \cmark & \xmark & 1{,}645 & 9.0$^\S$ \\
MMRC~\citep{xue_mmrc_2025}                      & \xmark & \cmark & \cmark & \xmark & \xmark & \cmark & \xmark & 5{,}120 & 12.9 \\
Mem-Gallery~\citep{bei_mem-gallery_2026}        & \cmark & \cmark & \cmark & \xmark & \xmark & \cmark & \cmark$^\dagger$ & 1{,}000 & 16.5 \\
MemLens~\citep{ren_memlens_2026}                & \cmark & \cmark & \cmark & \xmark & \xmark & \cmark & \xmark & 789 & 14--93$^\P$ \\
\midrule
ATM-Bench~\citep{mei_according_2026}            & \cmark & \cmark & \cmark & \cmark & \xmark & \cmark & \cmark & 1{,}069 & ---$^\ddagger$ \\
\midrule
MM-NIAH~\citep{wang_needle_2024}                & \xmark & \cmark & \cmark & \xmark & \xmark & \xmark & \xmark & 12{,}000 & --- \\
MRAG-Bench~\citep{hu_mrag-bench_2025}                 & \xmark & \cmark & \cmark & \xmark & \xmark & \xmark & \xmark & 16{,}130 & --- \\
\midrule
\textbf{MOSAIC (Ours)} & \cmark & \cmark & \cmark & \cmark & \cmark & \cmark & \cmark & \textbf{923} & \textbf{17.5} \\
\bottomrule
\end{tabular}
\caption{Comparison with prior memory benchmarks. \textbf{Cross-Mod.}: answering requires \emph{chaining} across modality boundaries. $^\dagger$Text--image co-reference only. $^\ddagger$Multi-source personal items (not dialogue sessions); avg.\ 1.6 evidence items/query. $^\S$110 dialogues, avg.\ 9 Q\&A turns. $^\P$Scales with 32K--256K context variants. Artefact counts: Table~\ref{tab:dataset-stats}.}
\label{tab:benchmark-comparison}
\end{table*}

Multi-hop text QA~\citep{yang_hotpotqa_2018, trivedi_musique_2022} likewise lacks this multimodal, temporal setting, and coupled system comparisons obscure \emph{where} evidence is lost. We therefore cast the pipeline as four stages---\textbf{write}, \textbf{retrieve}, \textbf{deliver}, and \textbf{read} (Figure~\ref{fig:overview})---and diagnose them with matched interventions under a frozen reader, prompt, budget, and judge.

Each stage fails in its own way, motivating four research questions:
\begin{itemize}[leftmargin=*,itemsep=0pt,topsep=2pt]
\item \textbf{RQ1 (Write).} Does the stored representation keep the spatial,
temporal, and prosodic detail the answer will later depend on?
\item \textbf{RQ2 (Retrieve \& Deliver).} Is every required unit found,
\emph{and} does it survive packing into the reader's request? Media caps can
drop even top-ranked artefacts, so the two are measured separately.
\item \textbf{RQ3 (Read).} Given the complete delivered chain, can the reader
integrate it across modalities and sessions?
\item \textbf{RQ4 (Abstain).} Does the system abstain precisely when
required evidence is absent---and only then?
\end{itemize}

MOSAIC has \textbf{923} questions in 11--24-session haystacks with
construction-targeted modality gaps (\S\ref{sec:modality-gaps}). Headline
findings: textualization costs $\sim$9 points (spatial-heavy); retrieval costs
$\sim$8 points and packing $\sim$4 at production budgets (far more under tight
media caps); $\sim$15 points remain under gold symbolic facts; abstention
remains weak on these controls. Under the matched protocol, each paired drop is
associated with the intervened stage; within-item carrier contrasts further
localize signal-specific loss, with the strongest causal validation for spatial
information and additional selector-level validation for annotated prosody
(\S\ref{sec:results:modality}).

\paragraph{Contributions.}
\begin{itemize}
\item \textbf{MOSAIC}, 923 persona-grounded omni-modal memory questions
with construction-targeted spatial, temporal, and prosodic signals (strongest
causal validation for spatial information; selector-level validation for
annotated prosody), cross-modal multi-hop chains, and 139 refusal controls.
\item \textbf{A matched four-stage diagnostic protocol} separating
\textsc{write}, \textsc{retrieve}, \textsc{deliver}, and \textsc{read},
with fixed-reader and native end-to-end evaluation tracks.
\item \textbf{A validated public benchmark pipeline}, including human
audits, media, filtering dispositions, manifests, and run configurations.
\end{itemize}

\section{Related Work}
\label{sec:related_work}

\paragraph{Conversational memory benchmarks.}
From single-session persona grounding~\citep{zhang_personalizing_2018} to multi-session suites, text-only benchmarks probe extraction, multi-session and temporal reasoning, knowledge updating, recall, and persona consistency~\citep{zhong_memorybank_2023, ou_dialogbench_2024, wu_longmemeval_2025, yu_memagent_2025, jiang_know_2025, he_madial-bench_2024, du_twinvoice_2025}. Multimodal extensions remain limited: LoCoMo images are often non-essential~\citep{maharana_evaluating_2024, bei_mem-gallery_2026}; Mem-Gallery and MemLens make images task-critical but stop at text+image and emphasise finding rather than cross-modal chaining~\citep{bei_mem-gallery_2026, ren_memlens_2026}; ATM-Bench aggregates personal multi-source evidence rather than sequential conversational chaining~\citep{mei_according_2026}; and MMRC, MMDU, MM-NIAH/MMNeedle, Mementos, and M3-Bench add image/video reasoning without multi-session dialogue structure~\citep{xue_mmrc_2025, liu_mmdu_2024, wang_needle_2024, wang_mementos_2024, zhou_m3-bench_2026}. MOSAIC is the first to require retrieving and chaining evidence across all four primary modalities in multi-session conversations (Table~\ref{tab:benchmark-comparison}).

\paragraph{Multi-hop and cross-modal reasoning.}
HotpotQA, MuSiQue, MultiModalQA, WebQA, and CRIT target compositional multi-hop QA over text and images~\citep{yang_hotpotqa_2018, trivedi_musique_2022, talmor_multimodalqa_2021, chang_webqa_2022, sung_crit_2026}; long-context suites such as BABILong, RULER, and MMLU-Reason chain facts in static corpora~\citep{kuratov_babilong_2024, hsieh_ruler_2024, tie_mmlu-reason_2025}. MOSAIC combines multi-hop reasoning with session-scattered multimodal retrieval.

\paragraph{Memory-augmented LLM systems.}
Memory architectures virtualize or structure long context~\citep{packer_memgpt_2024, chen_walking_2023, sarthi_raptor_2024, liu_memlong_2024}, integrate graphs with retrieval~\citep{guo_lightrag_2025, gutierrez_hipporag_2025, paul_gaama_2026}, or provide agentic/OS-inspired layers~\citep{xu_-mem_2025, kang_memory_2025, wang_o-mem_2025}. They lack a standard omni-modal benchmark for controlled comparison---the role MOSAIC fills.

\paragraph{RAG and memory diagnosis.}
Prior work localizes failures via claim-level RAG diagnostics~\citep{ru_ragchecker_2024}, tests of visual evidence use~\citep{hu_mrag-bench_2025}, controlled document pools~\citep{peng_unidoc-bench_2026}, AMemGym's write/read/utilization split~\citep{jiayang_amemgym_2026}, and MemFail's text-memory stages~\citep{garg_memfail_2026}; localization itself is not novel. We differ in \emph{how} stages are separated: instead of stressing a still-\emph{coupled} pipeline, MOSAIC makes each stage \emph{substitutable} under a frozen reader, prompt, packer, budget, and judge. This measures \textsc{deliver} apart from \textsc{retrieve} and attributes loss to carrier-level signals---to our knowledge the first substitutable four-stage protocol for omni-modal conversational memory.


\section{Preliminaries}
\label{sec:preliminaries}

Every MOSAIC item is a \emph{persona-grounded, multi-modal query} embedded in a chronological haystack of user--assistant sessions. Three orthogonal axes define it, each worked through end-to-end in Appendix~\ref{app:taxonomy}: the \textbf{question type} fixes the reasoning skeleton (\S\ref{sec:question-types}), the \textbf{modality gap} defines what each non-text modality must contribute (\S\ref{sec:modality-gaps}), and the \textbf{reasoning chain} composes \textsc{Narrower}, \textsc{Selector}, and \textsc{Answer} facts into a deterministic retrieval path (\S\ref{sec:chain}).

\subsection{Question Types}
\label{sec:question-types}

MOSAIC covers six types: single-session information extraction;
cross-session selection and bridging; knowledge updates; temporal-plan
disruptions; and refusal after evidence removal. Questions require one to
three facts; Appendix~\ref{app:taxonomy} defines each type, and
Appendix~\ref{app:dataset-stats} reports per-type counts.

\begin{table*}[t]
\centering
\small
\setlength{\tabcolsep}{4pt}
\begin{tabular}{@{}llp{0.54\textwidth}rr@{}}
\toprule
\textbf{Mod.} & \textbf{Challenge} & \textbf{Controlled realization} &
\textbf{Facts} & \textbf{Items} \\
\midrule
Image & Background detail & Put the target behind a more salient foreground subject. & 328 & 284 \\
Image & Fine-grained attribute & Require exact text, brand, shade, label, or a small count. & 388 & 333 \\
Image & Spatial configuration & Make relative position, rather than object presence, discriminative. & 194 & 185 \\
Video & Action ordering & Show two distinct completed actions in strict sequence without cuts. & 67 & 64 \\
Video & Repetition count & Repeat one action under a fixed camera, without a counter or overlay. & 64 & 56 \\
Video & State transition & Expose the start, change pattern, and end state. & 64 & 64 \\
Audio & Prosody & Hold words approximately neutral while happy or unhappy delivery selects an entity. & 450 & 399 \\
\bottomrule
\end{tabular}
\caption{The seven modality challenges and their overlapping coverage among the
784 answerable items. Counts are required-fact occurrences and distinct items,
not independent subsets.}
\label{tab:challenges}
\end{table*}

\subsection{Modality Gaps}
\label{sec:modality-gaps}

A \emph{modality gap} is information that survives only because its native
modality preserves it, while a text-only or caption-based agent would lose it.
Required non-text facts are assigned a gap strategy
(Appendix~\ref{app:taxonomy}): \textbf{image} (background detail, fine-grained
attribute, spatial configuration), \textbf{video} (action ordering, state
transition, repetition count), or \textbf{audio} (happy/unhappy prosody with
neutral wording over $\ge 2$ candidates); Appendix~\ref{app:dataset-stats}
breaks down tag coverage. Audio prosody is the most frequent
annotated strategy (450 facts / 399 of 784 answerable items), so Caption-Text /
ASR write paths lose the tone discriminator by design---a pipeline exposure,
not a gold-delivery perception cost (\S\ref{sec:results:modality}).

\paragraph{Challenge by construction.}
A media attachment is insufficient to establish multimodal reasoning,
since a caption or transcript may preserve the entire answer. MOSAIC therefore
assigns a \emph{challenge} before dialogue and media realization, specifying the
native relation that carries a fact together with constraints that keep that
relation out of the surrounding text; every answerable item contains at least
one required challenge-bearing fact (Table~\ref{tab:challenges}). Action
Repetition Count is representative: a fixed-camera clip repeats one action with
no cuts and no on-screen counter, and the dialogue says only ``a few times,'' so
a generic caption preserves the \emph{event} but loses the \emph{selector}
needed to continue the chain (per-challenge specifications:
Appendix~\ref{app:taxonomy}).

\subsection{Reasoning Chains and Fact Roles}
\label{sec:chain}

Every answerable item has a \emph{reasoning chain} of required facts with roles
\textsc{Narrower} (NAR; reduces candidates), \textsc{Selector} (SEL; identifies
an entity), and \textsc{Answer} (ANS; carries the final value), in the pattern
\(\text{\small NAR}^{*}\;\text{\small SEL}^{*}\;\text{\small ANS}\).
Valid chains are bridge-entity-linked and deletion-tested
(\S\ref{sec:pipeline:validation}): removing any required fact makes the
question unanswerable, and audio/video discriminators choose among
\(\ge 2\) same-type entities. A chain is \textbf{cross-modal} when required
facts span at least two modalities; gap-bearing detail usually sits in a
\textsc{Narrower} or \textsc{Selector}, and the \textsc{Answer}'s final
\emph{value} is always a short text-expressible string (enabling judge
scoring), though the fact carrying it may itself be multimodal (full
constraints and templates: Appendix~\ref{app:taxonomy}).

\section{The MOSAIC Generation Pipeline}
\label{sec:pipeline}
\label{sec:pipeline:chain}
\label{sec:pipeline:validation}
\label{sec:pipeline:sessions}
\label{sec:pipeline:stage2}
\label{sec:benchmark-stats}

MOSAIC is produced by a fully automated, config-driven four-stage pipeline
(Figure~\ref{fig:pipeline}; details in Appendix~\ref{app:pipeline}). Each item
is grounded in a Nemotron-Personas persona~\citep{nvidia/Nemotron-Personas-USA}.
An LLM designs a caption-distinct primary entity, then builds the chain
\textbf{question-first} under the bridge-entity and entity-pair constraints
(\S\ref{sec:chain}). A validate-and-repair loop combines programmatic checks
with a per-fact deletion test against twelve failure modes; unrepaired chains
are discarded, and \textsc{Answer Refusal} items strip required facts after
auditing. Surviving facts become chronological evidence and filler sessions
under leak guards. Stage~4 synthesises media (FLUX-2-pro / Kling--Seedance /
Qwen3-TTS). Filtering then yields the exact lineage
$1{,}050\!\rightarrow\!1{,}004\!\rightarrow\!879$ (automated) before a
three-annotator audit of 100 items (94.3\% well-posed) rescues 44 probe false
negatives, giving \textbf{MOSAIC-Eval}: \textbf{923} items (784 answerable +
139 refusal). Reweighting that audit with a 100-item probe-YES follow-up
estimates $\approx$97.8\% gold-correct and $\approx$96.6\% evidence-sufficient
on the 784 answerable releases (Appendix~\ref{app:human-eval}); the 784/864
(90.7\%) figure remains a \emph{filtering retention rate}, not a human
answerability ceiling (Appendix~\ref{app:filtering}).

The release spans 70 personas (11--24 sessions, mean 17.5), 10{,}765 multimodal
artefacts (Table~\ref{tab:dataset-stats}), and $\sim$18.8M tokens
($\sim$269K/persona). Among answerable items, 85.2\% need $\ge$2 facts and
75.0\% need $\ge$2 modalities; required evidence appears in image, audio,
video, and text for 99.5\%, 54.1\%, 23.5\%, and 35.3\% of items (overlapping),
so we report per-challenge strata rather than claiming uniform modality
balance; Appendix~\ref{app:dataset-stats} collects the remaining haystack
statistics.

\section{Diagnostic Evaluation Protocol}
\label{sec:experimental-setup}
\label{sec:evaluation}

To answer RQ1--RQ4 with matched stage contrasts, the protocol makes every stage
of the pipeline separately substitutable under a frozen reader, prompt, and
budget. For history $H$, question $q$, and delivery budget $B$, let
\begin{align}
M &= W(H), & R &= G(q,M), \nonumber\\
D &= P(R,B), & \hat{a} &= F(q,D), \label{eq:pipeline}
\end{align}
where $W$ writes memory, $G$ retrieves units, $P$ serializes and packs them,
and $F$ reads the delivered evidence. A retrieved artefact counts as
\emph{delivered} only if it is presented to the actual provider, a
distinction that turns out to matter a great deal
(\S\ref{sec:results:stages}).

\subsection{Matched Stage Decomposition}
\label{sec:protocol:stages}

\begin{table*}[t]
\centering
\small
\setlength{\tabcolsep}{5pt}
\begin{tabular}{@{}llrrrrrrl@{}}
\toprule
\textbf{Stage} & \textbf{Condition} & \textbf{Score} &
\textbf{Img} & \textbf{Vid} & \textbf{Aud} & \textbf{M-hop} &
\textbf{Chain} & \textbf{Diagnostic role} \\
\midrule
--- & Gold symbolic facts & 0.846 & 0.847 & 0.900 & 0.821 & 0.853 & 1.000 & Bounds residual read error \\
\textsc{write} & Gold native evidence & 0.783 & 0.785 & 0.729 & 0.826 & 0.783 & 1.000 & Native perception cost \\
\textsc{write} & Gold generic text & 0.690 & 0.693 & 0.625 & 0.718 & 0.682 & 1.000 & Textualization loss \\
\textsc{write} & Gold structured text & 0.704 & 0.705 & 0.652 & 0.742 & 0.702 & 1.000 & Avoidable share of it \\
\textsc{retrieve} & Retrieved native, all delivered & 0.708 & 0.709 & 0.631 & 0.741 & 0.696 & 0.675 & Retrieval loss \\
\textsc{deliver} & Retrieved native, capped & 0.670 & 0.671 & 0.576 & 0.696 & 0.652 & 0.491 & Packing loss \\
--- & End-to-end native RAG & 0.653 & 0.653 & 0.566 & 0.670 & 0.634 & 0.464 & Adds query rewrite \\
\midrule
\multicolumn{2}{@{}l}{\emph{Perception cost} (symbolic $-$ native, pts)} & 6.2 & 6.2 & 17.1 & $-0.5$ & 7.0 & --- & \\
\bottomrule
\end{tabular}
\caption{Matched \textsc{write}--\textsc{retrieve}--\textsc{deliver}--\textsc{read}
decomposition on the 784 answerable items with the fixed Gemini~3.5 reader
(Figure~\ref{fig:stage-ladder}). \textbf{Image}/\textbf{Video}/\textbf{Audio}
restrict to the 780/184/424 items whose chain needs that modality;
\textbf{Multi-hop} to the 668 needing $\ge$2 facts; \textbf{Chain} is
complete-chain delivery into the request. Paired CIs (and W/T/L for write
contrasts) are in Appendix~\ref{app:additional-results}.}
\label{tab:stage-results}
\vspace{-0.2in}
\end{table*}

We run seven matched conditions on the same 784 answerable QIDs
(Table~\ref{tab:stage-results}), changing only representation, retrieval, or
delivery under one reader, prompt, packer, and judge. These are
\emph{interventions} rather than an assumed ranking: native evidence may remain
perceptually difficult, and retrieved context may include useful evidence
outside the minimal annotated chain.

The main paired contrasts are symbolic versus native (perception cost),
structured versus generic text (avoidable write-time compression), gold native
versus retrieved-all-delivered (retrieval loss), and all-delivered versus
production-capped requests (delivery loss). Delivery is additionally traced as a
\emph{budget curve} over tighter packer allowances (top-10 units; top-5 with one
media item per modality). Residual errors under gold symbolic facts bound
answer integration and grading error.

\paragraph{Frozen configuration.}
Matched conditions use Gemini~3.5 as the fixed omni-modal reader (temperature~0;
extended thinking disabled), held out from dataset construction: MiMo-V2.5 is
the answerability probe, so using it as the primary reader would couple
selection with evaluation. The trade-off is disclosed---gold-native failures
mix perception error with bounded item noise
(\S\ref{sec:results:stages}). Retrieval replays a frozen top-50 ranked
manifest; the production budget delivers the top~20 units with at most three
items per modality under a 100\,MB media cap. Embedders, captioner/ASR,
structured writer (query-blind MiMo-V2.5 schemas), and judge (Qwen3.7-Plus)
are listed in Appendix~\ref{app:eval-setup}. MiMo-V2.5 as second reader on the
120-item pilot reproduces the gold and budgeted stage ladder
(Appendix~\ref{app:second-reader}); uncapped all-delivered remains Gemini-only
(MiMo timeouts under large multimodal payloads).

\subsection{Within-Item Modality Interventions}
\label{sec:protocol:modality}

Within an item we change one required carrier while holding its question,
evidence order, and every other hop fixed. Image variants are raw media,
OCR-only text, object-only text, and a spatial description; video variants are
raw video, ordered sparse frames, shuffled frames, and an ordered event/count
description; audio variants are raw audio, ASR, ASR plus a \emph{gold}
prosody tag from the construction annotation (an upper bound on any prosody
tagger), and ASR plus the \emph{swapped} gold tag (\texttt{happy}$\leftrightarrow$\texttt{unhappy}).
Image and video text variants use the same query- and answer-blind MiMo writer
as the structured-write condition.

Because every retained item was verified answerable from native evidence alone,
a text-over-raw gap indicates a perception failure rather than a missing
signal. Each contrast is restricted to the matching challenge stratum, and a
per-condition leak audit verifies that non-intervened hops are unchanged.

\subsection{Two Architecture Tracks}
\label{sec:protocol:tracks}

We separate two questions that end-to-end leaderboards conflate.
In \textbf{Track~A} (fixed reader), flat RAG over caption-text, omni-text, and
native indexes, plus A-MEM~\citep{xu_-mem_2025} and
MemoryOS~\citep{kang_memory_2025}, export retrieved context to one reader,
prompt, packer, budget, and judge---comparing the \emph{useful evidence each
backend produces}. In \textbf{Track~B} (native end-to-end), Naive RAG, repaired
ReAct~\citep{yao_react_2023}, A-MEM, and MemoryOS use their intended
controllers and readers, measuring \emph{deployable capability} descriptively
because multiple components change at once.

Three representations underlie both tracks
(Appendix~\ref{app:eval-setup}): \textbf{Caption-Text} (VL captioner + ASR),
\textbf{Omni-Text} (single omni LM captions that retain prosody Whisper
discards; retrieval still textual), and \textbf{Native-MM} (multimodal
embeddings + omni reader over raw artefacts). A-MEM and MemoryOS are
text-derived only; ReAct results in Table~\ref{tab:endtoend} are adaptive (up to 5 iterative search steps; mean 1.91 searches, median 1). A one-search parity gate is applied first---the scaffold is restricted to a single retrieval and must match the corresponding Naive RAG run's retrieval and delivery plumbing---and adaptive ReAct numbers are reported only after that gate passes.

\subsection{Metrics and Uncertainty}
\label{sec:protocol:metrics}

The primary answer metric is a 0--1 semantic judge score from Qwen3.7-Plus
(rubric in Appendix~\ref{app:eval-setup})~\citep{li_generation_2024}, reported
separately for the 784 answerable items, with correct-refusal and false-refusal
rates \emph{paired} so that an always-abstain system cannot appear calibrated.
Retrieval and delivery use required-fact and complete-chain Recall@$k$ over both
ranked units and units actually serialized into the request. Controlled
comparisons are paired by QID with item-bootstrap 95\% intervals and
wins/ties/losses; a persona-clustered bootstrap is a dependence check.
Reported intervals are per-contrast and unadjusted across the many paired
tests and strata, so exploratory challenge-level nulls should be read
cautiously. GLM-5.2 and Grok-4.5 evaluate each native end-to-end run
(Appendix~\ref{app:cross-judge}); Appendix~\ref{app:eval-setup} documents
manifests, media caps, and scoring; Appendix~\ref{app:prompts} reproduces the
prompts.

\section{Results and Analysis}
\label{sec:results}

All numbers come from one canonical paired run, and every stage claim uses only
paired conditions with identical QIDs, readers, prompts, packers, and judges. A
retrieved-but-dropped artefact counts as a \emph{delivery} failure rather than a
reading failure, and a write loss is claimed only where structured text recovers
an error made with generic text.

\subsection{Overall Stage Decomposition (RQ1--RQ3)}
\label{sec:results:stages}

\begin{figure*}[t]
\centering
\resizebox{0.88\textwidth}{!}{%
  \includegraphics{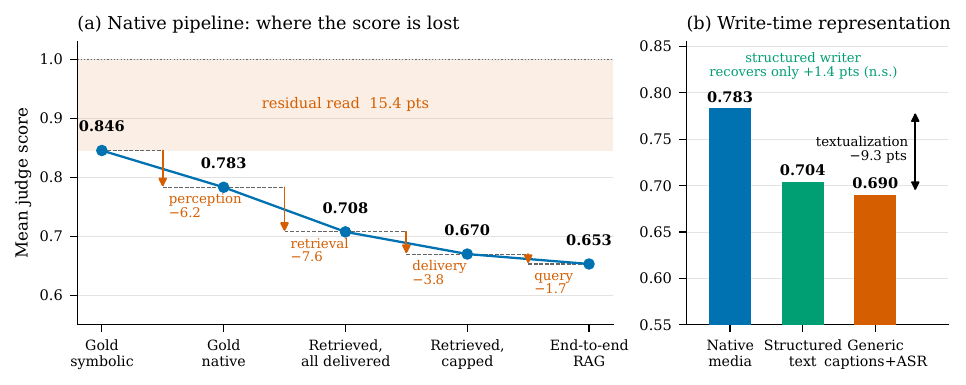}%
}
\caption{Decomposing the loss on the 784 answerable items under one fixed
reader, prompt, packer, and judge (Table~\ref{tab:stage-results}).
\textbf{(a)} Descending the native pipeline: each arrow is a paired drop
associated with the intervened stage under the matched protocol; the shaded
band is residual reading error under gold symbolic facts. \textbf{(b)}
Write-time representation only: generic captions/ASR cost 9.3 points; a
query-blind structured writer recovers a statistically insignificant 1.4.}
\label{fig:stage-ladder}
\end{figure*}

Table~\ref{tab:stage-results} and Figure~\ref{fig:stage-ladder} associate
each paired drop with the intervened stage under the matched protocol. Gold
symbolic facts score 0.846, so 15.4 points of residual reading/grading error
remain even with perfect evidence. Native gold evidence drops to 0.783
(6.2-point perception cost [3.2, 9.3]); generic captions/ASR fall further to
0.690 ($-9.3$ vs.\ native), and the structured writer recovers only $+1.4$
[$-0.9$, 3.7] to 0.704---a recovery indistinguishable from zero. Retrieval
costs 7.6 points (0.783$\to$0.708 all-delivered [5.2, 9.9]) and production
packing another 3.8 on identical ranked hits (0.708$\to$0.670 [1.5, 6.0]),
with complete-chain delivery falling from 0.675 to 0.491; end-to-end native
RAG reaches 0.653 ($+1.7$ vs.\ the original-query budgeted condition is not
material). Under this frozen reader and judge, reading is the largest single associated loss and the only
stage whose dominance is stable across pipelines; among the rest, retrieval leads for native
pipelines while write-time textualization leads for text-derived ones. These are matched-protocol associations, not universal causal bottlenecks: the 15.4-point gold-symbolic residual reflects reader, output-format, and judge error in addition to true reading failure.
Appendix~\ref{app:additional-results} gives paired CIs and write-contrast
W/T/L; Appendix~\ref{app:extended-stage} the per-type ladders.

\paragraph{Perception cost is modality-dependent.}
Video items score 0.900 with symbolic facts but only 0.729 with native video
(17.1-point perception cost), against 6.2 for image (0.847$\to$0.785) and none
for audio (0.821$\to$0.826). For this fixed reader, video perception is a
candidate bottleneck---not explained by retrieval or question difficulty alone.

\paragraph{Probe selection.}
Probe-pass ($n{=}740$) and human-rescued ($n{=}44$) items match under gold
symbolic facts (0.846 vs.\ 0.845) but differ by 27.8 points under gold native
evidence (0.799 vs.\ 0.520); absolute native scores are optimistic there, but
paired comparisons remain interpretable under the matched protocol.

\begin{figure}[t]
\centering
\includegraphics[width=\linewidth]{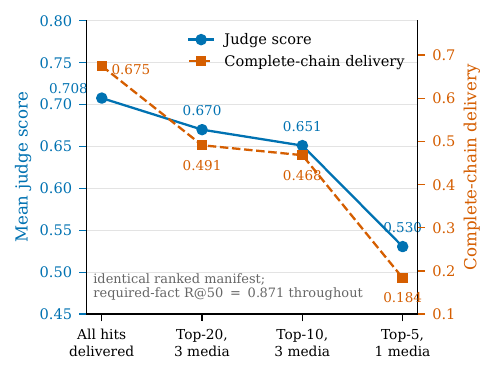}
\caption{Delivery-budget sweep over a single frozen top-50 ranked manifest.
Retrieval quality is constant (Recall@50 $=0.871$); only the packer's allowance
changes. One-media-per-modality packing collapses complete-chain delivery from
0.675 to 0.184.}
\label{fig:budget}
\vspace{-0.2in}
\end{figure}

\paragraph{Delivery is the stage a retrieval-terminating decomposition cannot
see.}
Figure~\ref{fig:budget} freezes the top-50 ranked manifest (required-fact
Recall@50 $=0.871$ throughout) and varies only the packer. Halving the unit
budget costs 5.7 points [3.3, 8.0]; one media item per modality \emph{jointly with the reduced top-$k$} costs 17.7
[14.5, 20.9] and collapses complete-chain delivery from 0.675 to 0.184
(required-fact coverage 0.871$\to$0.501). The 17.7-point drop therefore reflects the combined effect of the top-$k$ reduction and the one-media-per-modality cap together, not the media cap alone. Context construction, not ranking,
is the binding constraint once media budgets tighten (numeric sweep in
Appendix~\ref{app:budget}).

\subsection{Representation Loss by Modality (RQ1)}
\label{sec:results:modality}

\begin{figure}[t]
\centering
\includegraphics[width=\linewidth]{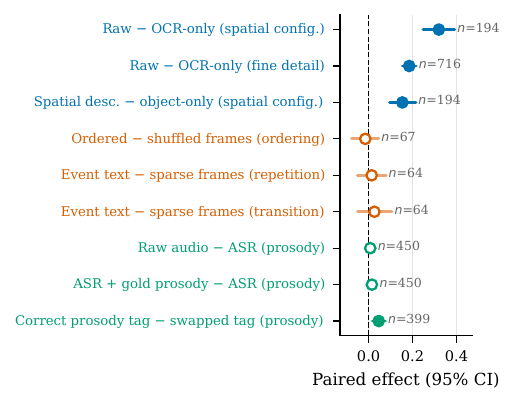}
\caption{Within-item paired effects on the governing challenge stratum
(item-bootstrap 95\% intervals). Filled markers exclude zero. Spatial structure
is the signal for which no tested text surrogate is adequate.}
\label{fig:modality}
\vspace{-0.2in}
\end{figure}

Figure~\ref{fig:modality} shows the paired effects on each governing challenge
stratum; Appendix~\ref{app:modality-interventions} tabulates exact estimates
and W/T/L.
\paragraph{Visual signals are heavily compressed by textualization.}
On fine-detail and background items, OCR-only text loses 18.5 points to raw
images [15.3, 21.6], and on spatial-configuration items raw images beat
OCR-only text by 32.0 points. What is lost is specifically \emph{relational}: on
those same spatial-configuration items, an explicit spatial description recovers
15.4 points over object-only text [9.5, 21.5], closing about a third of the
43.3-point gap between object-only text and raw pixels. Naming the objects
present is therefore not enough; the arrangement carries the fact, and a writer
that is told to preserve arrangement recovers a substantial part of it---but
only a part.

\paragraph{Temporal and prosodic signals show the opposite profile.}
Ordered sparse frames do not beat shuffled frames on action-ordering items
($-1.5$ points [$-7.8$, 4.6]); event/count text matches sparse frames on
repetition and state-transition items ($+1.4$ [$-5.2$, 8.0] and $+2.7$ [$-5.0$,
10.6]); \emph{raw audio} beats ASR by only 0.8 [$-1.5$, 3.0]; and a \emph{gold}
prosody tag adds only 1.5 [$-0.8$, 3.8] (394/450 ties). Under gold delivery
these are \emph{ceilings} on acoustic recovery once the transcript is in hand,
not evidence that the gap is decorative. On the same 399 items, Native-MM substantially outperforms Caption-Text
end-to-end (Appendix~\ref{app:gap-strategy}), a write/retrieve exposure
that gold delivery factors out. Video
strata remain small ($n{=}56$--$64$ unique QIDs); at 80\% power the MDEs are
$8.9$--$11.2$ points (Appendix~\ref{app:power}), so the nonsignificant video
contrasts do not establish lossless temporal compression.
Over 399 unique prosody QIDs, replacing the correct gold prosody tag with its
opposite reduced score by 4.6 points [1.7, 7.7; W/T/L 46/330/23]
(Appendix~\ref{app:modality-interventions}, Figure~\ref{fig:modality}): the
annotated selector is causally load-bearing under gold delivery, even though
raw audio versus ASR remains inconclusive.

\subsection{Fixed-Reader Comparison (RQ2)}
\label{sec:results:architecture}

\begin{table}[t]
\centering
\small
\setlength{\tabcolsep}{3.5pt}
\begin{tabular}{@{}lrr@{}}
\toprule
\textbf{Backend / representation} & \textbf{Chain R@$k$} & \textbf{Score} \\
\midrule
Flat RAG, caption text & 0.698 & 0.522 \\
Flat RAG, omni text & 0.705 & 0.500 \\
Flat RAG, native multimodal & 0.466 & 0.641 \\
A-MEM, caption text & \textbf{0.787} & \textbf{0.684} \\
A-MEM, omni text & 0.702 & 0.664 \\
MemoryOS, caption text & n/o & 0.268 \\
MemoryOS, omni text & n/o & 0.279 \\
\bottomrule
\end{tabular}
\caption{Track~A under the fixed Gemini~3.5 reader on 784 answerable QIDs.
Chain R@$k$ denotes complete-chain delivery; MemoryOS coverage is unobservable.
Paired effects: Appendix~\ref{app:track-a}.}
\label{tab:track-a}
\vspace{-0.2in}
\end{table}

With the reader fixed (Table~\ref{tab:track-a}), A-MEM beats same-representation
flat RAG by 16.2 points on caption text [13.1, 19.3; W/T/L 233/464/87] and by
16.4 on omni text [13.2, 19.4]; A-MEM caption (0.684) also outscores flat RAG
over \emph{native} multimodal evidence (0.641). Chain coverage explains why:
A-MEM delivers the complete required chain on 0.787 of items vs.\ 0.698 (flat
caption) and only 0.466 (native index), whose media caps evict chain members.
MemoryOS lags at 0.268/0.279 (25.5 / 22.1 points behind flat RAG), consistent
with Track~B over-abstention.
Memory organization therefore helps when it improves \emph{complete-chain
delivery} under a fixed reader and budget; a native index can still lose if
packing drops required media before the reader sees them. Backends that keep
every required hop in the request win even with a text-only store; native
perception cannot rescue a truncated chain. All 9{,}338 retrieved-but-dropped
artefacts hit the media-count cap (Appendix~\ref{app:drop-reasons}).

\subsection{End-to-End Systems \& Abstention (RQ4)}
\label{sec:results:endtoend}

\begin{figure}[t]
\centering
\includegraphics[width=\linewidth]{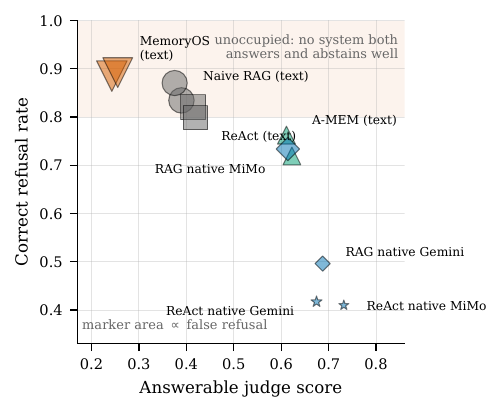}
\caption{Answering versus abstention for all twelve Track~B systems. Marker
area is proportional to the false-refusal rate. The shaded upper region is
unoccupied. Full scores: Appendix~\ref{app:endtoend},
Table~\ref{tab:endtoend}.}
\label{fig:calibration}
\vspace{-0.2in}
\end{figure}

Figure~\ref{fig:calibration} exposes a sharp answer--abstention trade-off.
Native ReAct with MiMo-V2.5 leads on answerable items (0.732) but correctly
refuses only 0.410, whereas MemoryOS exceeds 0.88 correct refusal while scoring
below 0.26 on answerable items and falsely refusing over 0.61. A-MEM caption
text is the closest to balanced, but no system occupies the upper-right.
Because refusal controls retain question-form cues
(Appendix~\ref{app:refusal-structure}), RQ4 diagnoses abstention under this
control set rather than form-blind calibration. 

\section{Conclusion}
\label{sec:conclusion}

MOSAIC separates omni-modal memory failures into \textsc{write},
\textsc{retrieve}, \textsc{deliver}, and \textsc{read}. Under the matched
protocol and fixed reader, reading accounts for the largest residual loss
under gold facts; retrieval and delivery constrain native pipelines, textualization text-derived ones. These are protocol associations, not universal bottlenecks. Spatial structure has the
clearest causal validation and no adequate text surrogate; annotated prosody
tags show selector-level sensitivity, while temporal and raw-audio versus ASR
contrasts remain inconclusive. No system both answers and refuses reliably:
progress requires preserving modality-specific signal, delivering complete
chains, and reading them faithfully.

\section{Limitations}
\label{sec:limitations}

MOSAIC's personas, evidence sessions, and media are synthetic; evaluation on
organically collected multimodal histories remains future work. The primary
reader is held out from filtering, and stage contrasts are paired, but stage
labels are matched-protocol associations rather than fully isolated mechanisms.
Video nulls are underpowered (Appendix~\ref{app:power}). The raw-audio versus
ASR gap is inconclusive for this reader, and textualized audio carriers often
retain paralinguistic lexical cues while other gold hops already expose the
answer value---so gold-delivery audio nulls do not establish acoustic recovery
(Appendix~\ref{app:extended-limitations}). Swapped gold prosody tags
nonetheless show selector-level sensitivity
(Appendix~\ref{app:modality-interventions}). RQ4 uses length-matched refusal
controls with residual form cues (Appendix~\ref{app:refusal-structure}).
Further caveats on modality coverage and backend scope are in
Appendix~\ref{app:extended-limitations}.

\section{Ethical Considerations}
\label{sec:ethics}

Personas, evidence sessions, and all media artefacts in MOSAIC are synthetic; we
release no private user logs collected by the authors. Filler \emph{dialogue
text} is sampled from LongMemEval's public filler-session pool
(Appendix~\ref{app:filler}): UltraChat under MIT and ShareGPT under Apache~2.0
as stated by LongMemEval; multimodal attachments on those fillers are synthetic
injections from our pipeline. Stage-localized evaluation is intended as a
constructive tool: deployers can see \emph{where} evidence was dropped rather
than only that a system scored poorly. The human audit used short voluntary
academic annotation (Appendix~\ref{app:human-eval}). Real deployments of
omni-modal memory should of course obtain user consent and provide retention
controls, concerns outside the scope of this benchmark release. AI assistants
aided coding and drafting.

\bibliography{custom}

\clearpage
\appendix
\raggedbottom

\section{Full Taxonomy Details}
\label{app:taxonomy}

\subsection{Question Type Definitions and Examples}

Table~\ref{tab:question-types} summarises hop and session requirements for the
six types introduced in \S\ref{sec:question-types}; Table~\ref{tab:taxonomy-types}
gives full definitions and examples from the 923-item evaluation set.

\begin{table}[!ht]
\centering
\small
\setlength{\tabcolsep}{3pt}
\begin{tabular}{@{}lccl@{}}
\toprule
\textbf{Type} & \textbf{Hops} & \textbf{Sess.} & \textbf{Reasoning} \\
\midrule
Info.\ Extraction      & 1--2 & 1        & Direct recall \\
\textsc{Msr-Selection} & 2--3 & $\geq$2  & Cross-sess.\ select \\
\textsc{Msr-Bridge}    & 3    & $\geq$3  & Sequential chain \\
Knowledge Update       & 2    & $\geq$2  & Old$\rightarrow$new \\
Temporal Reasoning     & 2--3 & $\geq$2  & Plan$\rightarrow$disrupt \\
Answer Refusal         & 1--2 & 1        & Abstention \\
\bottomrule
\end{tabular}
\caption{The six MOSAIC question types. \textbf{Hops} = required facts in the
reasoning chain; \textbf{Sess.} = evidence sessions that host those facts.}
\label{tab:question-types}
\end{table}

\begin{table*}[!ht]
\centering
\small
\setlength{\tabcolsep}{3pt}
\begin{tabular}{@{}lp{0.35\textwidth}p{0.40\textwidth}@{}}
\toprule
\textbf{Type} & \textbf{Definition} & \textbf{Example} \\
\midrule
Information Extraction & Single-session factual recall. Disambiguates among $\ge 2$ same-category items via a discriminator (tone, sequence, count, temporal hint). 1--2 hops. & \textit{``In the photo of my fondue set with the colour-coded forks, what brand is stamped on it?''} \\
\midrule
MSR Selection & Cross-session entity binding: entities of the same category appear in different sessions; a discriminator selects the target. 2--3 hops, $\ge 2$ sessions. & \textit{``What background detail is visible in the photo of the tomato seedlings I sounded happy about?''} \\
\midrule
MSR Bridge & Sequential chain: entity $e_0$ in $s_0$ identifies $e_1$ in $s_1$, which identifies $e_2$ in $s_2$; the answer is an attribute of $e_2$. 3 hops, $\ge 3$ sessions. & \textit{``What is the packing method from the carver in the video where I first hollowed then inlaid, in the town on the sign in my chess set photo?''} \\
\midrule
Knowledge Update & Evolving attribute tracking (old $\rightarrow$ new state) across sessions. 2 hops, $\ge 2$ sessions. & \textit{``In the two photos, what brand did my chore coat change to?'' (Carhartt $\rightarrow$ Dickies)} \\
\midrule
Temporal Reasoning & Time-order reasoning with a causal disruption: an old plan is altered; the answer depends on the new outcome. 2--3 hops, $\ge 2$ sessions. & \textit{``After I sounded unhappy about the tour cancellation, what was the shade name of the replacement tile I photographed?''} \\
\midrule
Answer Refusal & Evidence-stripped unanswerable calibration check. The question appears answerable on its surface, but the required evidence has been deliberately removed. 1--2 hops, 1 session. & \textit{``In the photo of my Kierkegaard book on the bench by the foggy harbour, what stand-up comedy detail is in the background?'' (Expected: ``I don't know.'')} \\
\bottomrule
\end{tabular}
\caption{The six MOSAIC question types with definitions and concrete examples.}
\label{tab:taxonomy-types}
\end{table*}

\subsection{Modality Gap Strategies}

Required non-text facts are assigned a gap strategy (1{,}555/1{,}582 carry a
named tag in the release; see Table~\ref{tab:gap-strategies}). A modality gap
is information that survives only because its native modality preserves
it---a text-only or caption-based agent would lose it. The gaps are not
arbitrary: each probes a known encoder failure mode so that recovering the
answer requires the native channel that mechanism preserves.

\paragraph{Encoder motivation (deferred from \S\ref{sec:modality-gaps}).}
\textbf{Image gaps (spatial)} target detail that patch-based visual encoders compress away under token-pooling and resolution limits, and that captioners---which describe the salient foreground---omit. \textbf{Video gaps (temporal)} target what sparse frame-sampling and temporal pooling destroy: a caption of ``assembling a shelf'' retains neither order nor count. \textbf{Audio gaps (prosodic)} target the lexical--acoustic split that ASR collapses; prosody is the signal most completely erased by textualization.

\paragraph{Image Gaps (Spatial).}
\begin{itemize}[leftmargin=*,itemsep=1pt]
\item \textbf{Background detail:} The answer resides in the periphery or background; the foreground depicts an unrelated object. \textit{Example: a photo of a chess set where the answer is the lighthouse visible through the window behind it.}
\item \textbf{Fine-grained attribute:} The answer requires reading text (brand names, prices, serial numbers, book titles) or counting exact quantities visible only in the image. \textit{Example: the brand name ``Lorraine'' stamped on the base of a fondue pot, never mentioned in chat text.}
\item \textbf{Spatial configuration:} The answer depends on the relative arrangement or position of items. \textit{Example: ``the left pole leaning higher than the right'' in a photo of hiking poles against a fence.}
\end{itemize}

\paragraph{Video Gaps (Temporal).}
\begin{itemize}[leftmargin=*,itemsep=1pt]
\item \textbf{Action ordering:} Which of two distinct actions came first vs.\ second. \textit{Example: a video of a shelf being assembled, where the model must determine whether the photo was placed before or after the candle was lit.}
\item \textbf{State transition:} How a change unfolded---gradual, sudden, uneven, patchy. \textit{Example: a barn renovation video showing transformation stages across multiple weeks.}
\item \textbf{Action repetition count:} How many times an action occurred, banded as 1--2, 4--5, or 8+ to absorb video-generation variance. \textit{Example: counting how many times a molinillo is spun in a hot chocolate preparation video.}
\end{itemize}

\paragraph{Audio Gap (Prosodic).}
\begin{itemize}[leftmargin=*,itemsep=1pt]
\item \textbf{Audio prosody (happy/unhappy):} Emotional tone conveys preference, while the transcript uses deliberately neutral words naming $\ge 2$ candidates. \textit{Example: a voice message saying ``I tried the brass-and-glass telegraph'' in a happy tone, while another message mentions a different telegraph neutrally.}
\end{itemize}

\subsection{Fact Roles and Chain Structure}

Every answerable MOSAIC item is built around a reasoning chain---an ordered sequence of \emph{required facts}. Each fact has a role:
\begin{itemize}[leftmargin=*,itemsep=1pt]
\item \textbf{\textsc{Narrower} (NAR):} Reduces the candidate set (e.g., ``I prefer vintage items'').
\item \textbf{\textsc{Selector} (SEL):} Identifies a unique entity (e.g., ``the brass-and-glass telegraph I sounded happy about'').
\item \textbf{\textsc{Answer} (ANS):} Carries the final value (e.g., ``8:15 AM'').
\end{itemize}
A chain follows the pattern $\text{\small NAR}^{*}\;\text{\small SEL}^{*}\;\text{\small ANS}$. Every valid chain satisfies four constraints defined in \S\ref{sec:chain}. A chain is \textbf{cross-modal} when its required facts span at least two modalities. The gap-bearing detail typically resides in a \textsc{Narrower} or \textsc{Selector}. The \textsc{Answer}'s final \emph{value} is always a short text-expressible string (a name, price, count, or date), but the fact that carries it may itself be multimodal---e.g., a brand name readable only in an image, or a preference conveyed only by prosody.

\section{Generation Pipeline Details}
\label{app:pipeline}

\begin{figure*}[t]
    \centering
    \includegraphics[width=\linewidth]{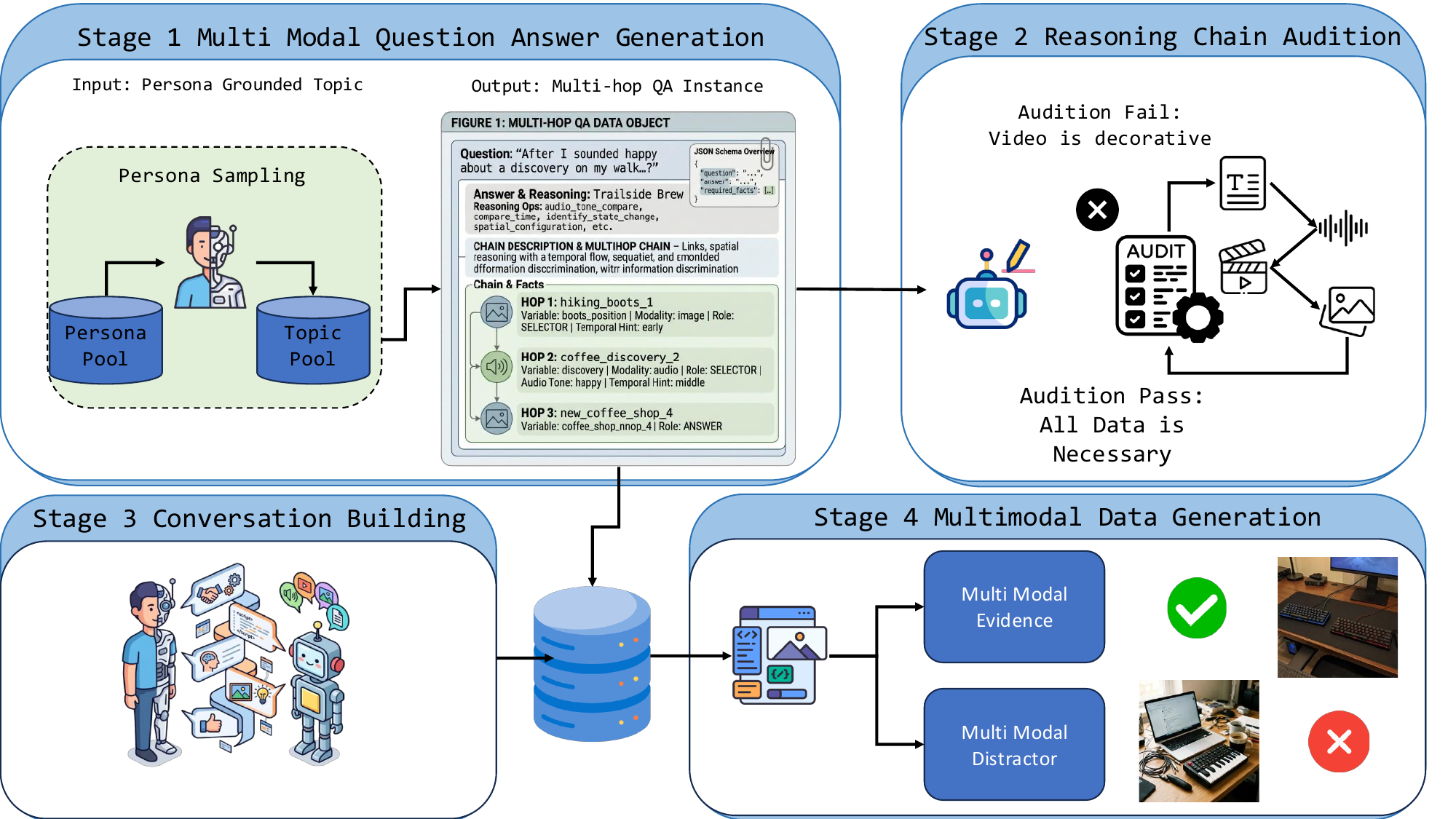}
    \caption{The MOSAIC generation pipeline. \textbf{Stage~1} samples a persona and designs a multi-hop reasoning chain; \textbf{Stage~2} audits it via a per-fact deletion test against twelve failure modes; \textbf{Stage~3} realises it as a timestamped multi-session conversation with persona-conditioned filler; \textbf{Stage~4} synthesises the media assets, followed by structural and LLM-based quality filtering.}
    \label{fig:pipeline}
\end{figure*}

\subsection{Stages~1--3 in Detail}
\label{app:pipeline:stages13}

Each item is grounded in a persona from Nemotron-Personas. A round-robin sampler assigns gap strategies and audio tones for uniform coverage, and a single LLM call designs one \emph{caption-distinct} primary entity per question (at most two per category, differing in $\ge 2$ of style, colour, material, setting, or condition). The LLM then designs the chain \textbf{question-first}---write the question, choose the answer, then construct the chain $\mathcal{C}$ and its facts under the \emph{bridge-entity} and \emph{entity-pair} constraints (\S\ref{sec:chain}). Every chain enters a \textbf{validate-and-repair loop} (up to three iterations): a programmatic pass enforces hard structural constraints, and an LLM judge runs a \textbf{per-fact deletion test} against twelve named failure modes (e.g.\ \textsc{Lexical-Leak}, \textsc{Bridge-Integrity}, \textsc{Shortcut}); chains unrepaired after three iterations are discarded, and \textsc{Answer Refusal} items have their required facts stripped after auditing. Surviving facts are realised as natural turns in a chronological haystack of evidence and persona-conditioned filler sessions, under a temporal-leak guard and a forbidden-value list that keep early or non-answer turns from leaking the answer.

\subsection{Stage~4 Filtering Lineage and Release Statistics}
\label{app:pipeline:stage4-stats}

Stage~4 synthesises media---images via FLUX-2-pro (GPT-image-2 for text-rich gaps), video via Kling/Seedance, and audio via Qwen3-TTS with persona-aware voices---after which a filtering cascade reduces the 1{,}050 raw items to 1{,}004 structurally and semantically valid candidates. Of the 864 answerable candidates, an omni answerability probe accepts 745; a second automated quality intersection retains 740 of these plus 139 of the 140 \textsc{Answer Refusal} candidates, producing an 879-item automated set. A three-annotator human audit of 100 sampled items (94.3\% well-posed) then rescues 44 probe false negatives, yielding exactly \textbf{923} unique items (784 answerable + 139 refusal). We report this as an exact lineage rather than reconstructing it from rounded pass rates; the 784/864 (90.7\%) figure is a filtering retention rate, not a human answerability ceiling (full cascade in Appendix~\ref{app:filtering}; audit in Appendix~\ref{app:human-eval}).

MOSAIC-Eval spans 70 personas (11--24 sessions, mean 17.5), 4{,}573 annotated facts of which 1{,}860 are required by answerable items, 10{,}765 multimodal artefacts, and an $\sim$18.8M-token haystack ($\sim$269K tokens per persona, $1.7\times$ LongMemEval's budget in denser sessions). Among answerable items, 668/784 (85.2\%) require at least two facts and 588/784 (75.0\%) require at least two modalities. Required evidence appears in image, audio, video, and text for 780 (99.5\%), 424 (54.1\%), 184 (23.5\%), and 277 (35.3\%) items respectively; these incidence rates overlap, so we report per-challenge strata throughout. The remaining distribution tables appear in Appendix~\ref{app:dataset-stats}.

\subsection{Prompt Templates}

The pipeline uses a modular set of LLM prompt templates, each targeting a specific generation subtask:

\begin{itemize}[leftmargin=*,itemsep=2pt]
\item \textbf{Chain design:} Given a persona, question type, and assigned modalities, the LLM designs the reasoning chain question-first: write the question, choose the answer, then construct the chain $\mathcal{C}$ and its facts. Type-specific templates enforce the required hop count and session structure.
\item \textbf{Fact extraction:} After chain design, the LLM extracts canonical fact statements from the reasoning chain, assigning each fact a role (\textsc{Narrower}, \textsc{Selector}, \textsc{Answer}), a modality, and a gap strategy.
\item \textbf{Session planning:} Facts are distributed across a chronological timeline of evidence and filler sessions. Periods and session groups are assigned programmatically; the LLM determines the natural language realisation of each fact within its assigned session.
\item \textbf{Validation and repair:} A dedicated validation prompt runs a per-fact deletion test and checks for twelve named failure modes (see below). Failures trigger minimal repairs (1--3 targeted changes); items unrepaired after three iterations are discarded.
\item \textbf{Session building:} An LLM rewrites each canonical fact as a natural conversation turn under two guards: a temporal-leak guard restricts facts to those with $\texttt{hop\_index} \le$ the current turn, and a forbidden-value list prevents \textsc{Narrower} and \textsc{Selector} facts from mentioning the \textsc{Answer} value.
\end{itemize}

\subsection{Chain Templates}

Table~\ref{tab:chain-templates} lists the type-specific chain templates used during Stage~1 chain design (\S\ref{sec:pipeline:chain}).

\begin{table}[!ht]
\centering
\resizebox{0.9\linewidth}{!}{%
\begin{tabular}{@{}ll@{}}
\toprule
\textbf{Type} & \textbf{Chain template} \\
\midrule
Info.\ Extraction      & \textsc{Nar}\textsubscript{opt}~$\rightarrow$~\textsc{Sel}~$\rightarrow$~\textsc{Ans} \\
\textsc{Msr-Selection} & \textsc{Nar}\textsubscript{opt}~$\rightarrow$~\textsc{Sel}~$\rightarrow$~\textsc{Ans} (cross-sess.) \\
\textsc{Msr-Bridge}    & \textsc{Nar}\textsubscript{opt}~$\rightarrow$~\textsc{Sel}~$\rightarrow$~\textsc{Sel}~$\rightarrow$~\textsc{Ans} \\
Knowledge Update       & \textsc{Nar}\textsubscript{opt}~$\rightarrow$~\textsc{Sel}\textsubscript{old}~$\rightarrow$~\textsc{Ans}\textsubscript{new} \\
Temporal Reasoning     & \textsc{Nar}\textsubscript{opt}~$\rightarrow$~\textsc{Sel}\textsubscript{plan}~$\rightarrow$~\textsc{Sel}\textsubscript{disrupt}~$\rightarrow$~\textsc{Ans}\textsubscript{out.} \\
Answer Refusal         & inherits IE, evidence stripped \\
\bottomrule
\end{tabular}%
}
\caption{Chain templates. \textsc{Nar}~$=$ \textsc{Narrower}, \textsc{Sel}~$=$ \textsc{Selector}, \textsc{Ans}~$=$ \textsc{Answer}. ``opt'' denotes an optional narrowing step.}
\label{tab:chain-templates}
\end{table}

\subsection{Twelve Named Failure Modes}

Table~\ref{tab:failure-modes} catalogues the twelve failure modes checked during Stage~2 chain validation (\S\ref{sec:pipeline:validation}). Each mode targets a specific class of generation defect, detected by the LLM validation judge, triggering targeted repairs; chains unrepaired after three iterations are discarded.

\begin{table}[H]
\centering
\scriptsize
\setlength{\tabcolsep}{3pt}
\begin{tabular}{@{}lp{0.60\columnwidth}@{}}
\toprule
\textbf{Failure Mode} & \textbf{Definition} \\
\midrule
\textsc{Lexical-Leak}      & The question text paraphrases or contains the answer value. \\
\textsc{Bridge-Integrity}   & Adjacent facts in the chain do not share a concrete referent. \\
\textsc{Shortcut}           & The answer can be derived from text-only evidence, bypassing the intended modality gap. \\
\textsc{Entity-Mismatch}    & The entity described in the \textsc{Answer} fact does not match the entity selected by prior facts. \\
\textsc{Gap-Anchoring}      & The question text does not reference the gap-bearing detail, making it unclear which evidence to retrieve. \\
\textsc{Redundant-Gap}      & Two facts test the same modality gap, reducing diversity. \\
\textsc{Session-Overlap}    & Required facts from different hops are placed in the same session, collapsing the temporal structure. \\
\textsc{Tone-Collapse}      & Both audio candidates share the same prosodic tone, removing the discriminator. \\
\textsc{Period-Leak}        & A later-period fact leaks the answer value into an earlier period's context. \\
\textsc{Caption-Distinctness} & Primary entities in the same category are not sufficiently distinct ($<2$ of style, colour, material, setting, condition). \\
\textsc{Entity-Pair}        & An audio/video \textsc{Narrower} or \textsc{Selector} has no competing same-type entity, making the gap unnecessary. \\
\textsc{Count-Band}         & The action repetition count in a video question does not match the specified count band (1--2, 4--5, 8+). \\
\bottomrule
\end{tabular}
\caption{The twelve named failure modes checked during Stage~2 validation.}
\label{tab:failure-modes}
\end{table}



\subsection{Filler Session Generation}
\label{app:filler}

Filler \emph{dialogue text} is taken from LongMemEval's publicly released
custom-history filler pool~\citep{wu_longmemeval_2025}
(\texttt{data\_5\_filler\_sess.json}; 29{,}149 sessions):
UltraChat~\citep{ding_ultrachat_2023} (20{,}000 sessions; LLM-synthesized
dialogues; MIT; accessed via \url{https://github.com/thunlp/UltraChat}) and
ShareGPT~\citep{chiang_vicuna_2023} (9{,}149 sessions; user-shared ChatGPT
transcripts curated via LMSYS FastChat; Apache~2.0 as stated by
LongMemEval, accessed via \url{https://github.com/lm-sys/FastChat}). We do not
re-sample upstream dumps independently; session IDs retain the LongMemEval
\texttt{ultrachat\_*} / \texttt{sharegpt\_*} scheme. In the official release
configuration the existing-corpus ratio is 1.0 (no LLM-generated filler
dialogue). Multimodal attachments on fillers are injected separately by our
pipeline and are synthetic. Topic filtering prevents fillers from mentioning
evidence entities. Fillers are interleaved chronologically with evidence
sessions to create realistic temporal gaps.

\section{Evaluation Setup}
\label{app:eval-setup}

\subsection{System Architectures}

\begin{figure}[!ht]
    \centering
    \includegraphics[width=0.8\linewidth]{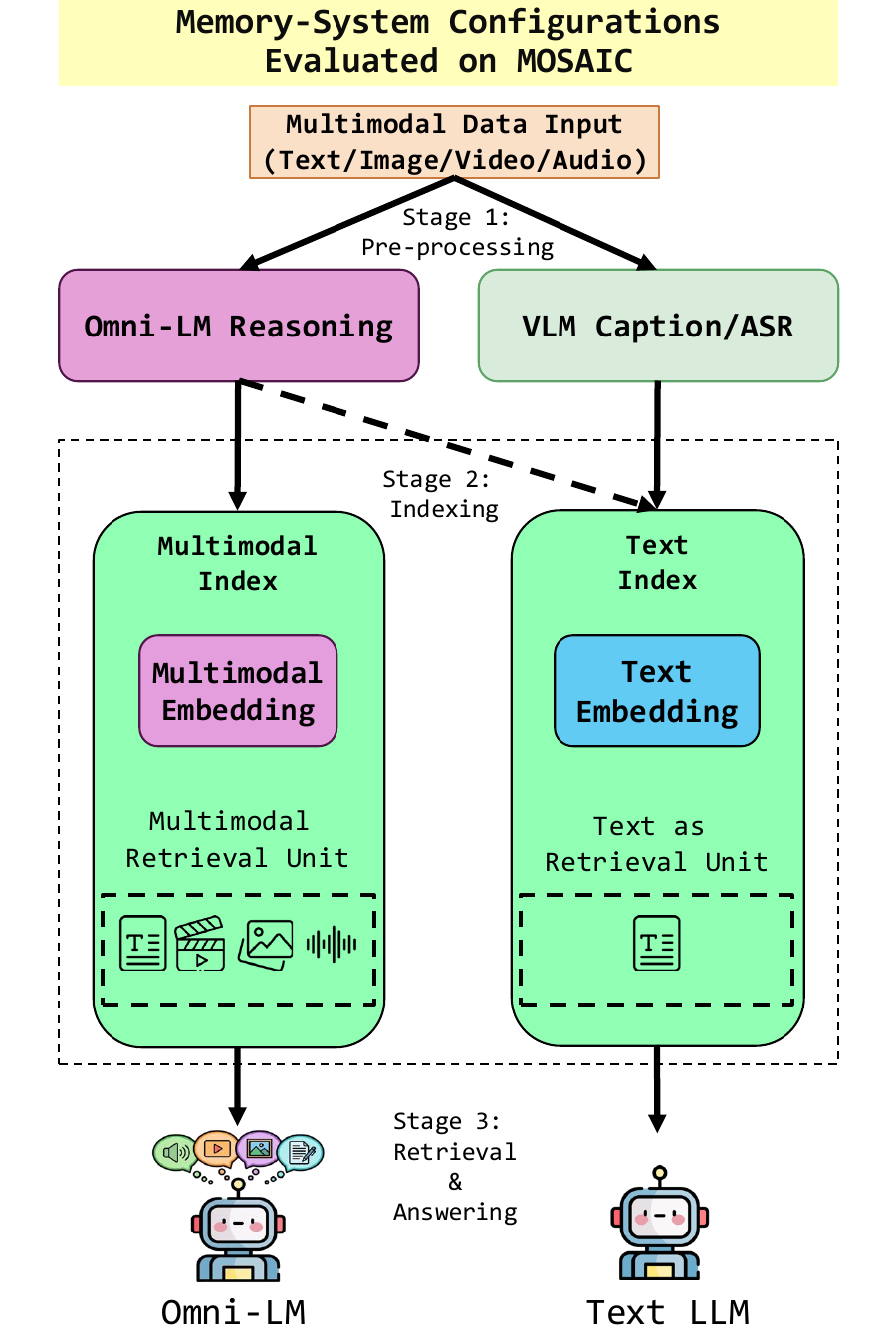}
    \caption{The three system architectures evaluated on MOSAIC: \textbf{Caption‑Text} (VL captioner + Whisper ASR, text index), \textbf{Omni‑Text} (single omni captioner preserving prosody, text index), and \textbf{Native‑MM} (multimodal embeddings over raw artefacts, omni‑modal reasoner).}
    \label{fig:architectures}
    \vspace{-0.1in}
\end{figure}

Three architectures differ in how multimodal evidence is preprocessed and indexed (see Figure~\ref{fig:architectures}):

\begin{itemize}[leftmargin=*,itemsep=2pt]
\item \textbf{Caption-Text (CT):} A VL captioner (Qwen3-VL-235B-A22B) and Whisper-large-v3-turbo convert all non-text evidence to captions. All captions are indexed as text chunks with a text embedding model (Qwen3-Embedding-0.6B). A text-only LLM reasoner retrieves and answers over captions.
\item \textbf{Omni-Text (OT):} A single omni-modal LM (MiMo-V2.5) replaces the cascade, generating captions that preserve audio prosody (Whisper discards it). Otherwise identical to Caption-Text: text index, text reasoner.
\item \textbf{Native-MM (NMM):} No textualization. Raw images, video frames, and audio are indexed with a multimodal embedding model (\texttt{jina-embeddings-v5-omni-small}). An omni-modal LM reasons directly over the retrieved artefacts (images, video, audio).
\end{itemize}

\subsection{Retrieval Back-Ends}
\begin{itemize}[leftmargin=*,itemsep=2pt]
\item \textbf{Naive RAG:} Single-pass retrieval with top-$k=20$, using Qwen3-Embedding-0.6B~\citep{zhang_qwen3_2025} for text indices and \texttt{jina-embeddings-v5-omni-small}~\citep{honicke_jina-embeddings-v5-omni_2026} for Native-MM. Query rewriting (LLM keyword extraction) before embedding. No time filter. Score threshold 0.0.
\item \textbf{ReAct-RAG:} Iterative retrieval (up to 5 steps) with a tool-use loop; the model decides when to search and when to answer. SEARCH/ANSWER marker format for Native-MM, function calling for text architectures. Search top-$k=20$, query rewriting enabled, duplicate query rejection. Every ReAct configuration must first pass a \emph{one-search parity gate}, in which the scaffold is restricted to a single retrieval call and must reproduce the corresponding Naive RAG run's retrieval and delivery plumbing; ReAct numbers are reported only after that gate matches.
\item \textbf{A-MEM~\citep{xu_-mem_2025}:} Agentic memory system that extracts structured memory entries from conversation history. Uses MiniLM-L6-v2~\citep{wang_minilm_2020} embeddings. Applied only to Caption-Text and Omni-Text, since it does not ingest native media.
\item \textbf{MemoryOS~\citep{kang_memory_2025}:} Memory operating system that organizes conversation history into retrievable memory blocks. Uses MiniLM-L6-v2 embeddings. Applied only to Caption-Text and Omni-Text. MemoryOS does not expose a ranked candidate list, so its chain-coverage cells are reported as not observable rather than zero.
\end{itemize}

\subsection{Reasoning Models}
\begin{itemize}[leftmargin=*,itemsep=2pt]
\item \textbf{Fixed reader (all matched conditions and Track~A):} Gemini~3.5, temperature 0, 1{,}024 output tokens, extended thinking disabled. Chosen because it played no role in dataset construction.
\item \textbf{Text reasoners (Track~B):} DeepSeek-v4-Flash for Naive RAG, ReAct, A-MEM, and MemoryOS over the text architectures. Temperature 0.
\item \textbf{Omni reasoners (Track~B, Native-MM only):} MiMo-V2.5 and Gemini~3.5. Temperature 0. For Native-MM ReAct, the omni model handles both the iterative search loop and the final answer.
\item \textbf{Write-time models:} generic text uses a Qwen3-VL-235B captioner~\citep{bai_qwen3-vl_2025} with Whisper-large-v3-turbo ASR~\citep{radford_robust_nodate}; structured text uses a query-blind MiMo-V2.5 writer with fixed per-modality schemas that preserve spatial relations, temporal order, counts, and prosody.
\end{itemize}

\subsection{Diagnostic Conditions}
\label{app:diagnostic-conditions}

Beyond the two architecture tracks, the matched decomposition
(\S\ref{sec:protocol:stages}) runs seven conditions on the same 784 answerable
QIDs under one reader, prompt, packer, budget, and judge:

\begin{itemize}[leftmargin=*,itemsep=2pt]
\item \textbf{Gold symbolic facts:} the annotated required facts as text, bounding residual read and grading error.
\item \textbf{Gold native evidence:} the same chain delivered as raw media, isolating native perception cost.
\item \textbf{Gold generic text:} the same chain as VL captions plus Whisper ASR, isolating write-time textualization loss.
\item \textbf{Gold structured text:} the same chain under the query-blind structured writer, isolating the \emph{avoidable} share of that loss.
\item \textbf{Retrieved native, all delivered:} every retrieved hit is serialized, isolating retrieval loss at fixed delivery.
\item \textbf{Retrieved native, production caps:} the identical ranked manifest under top-20 units, three media per modality, 100\,MB, isolating delivery loss at fixed ranking.
\item \textbf{End-to-end native RAG:} adds the frozen fixed-rewrite query ranking.
\end{itemize}

Delivery is additionally swept over two tighter budgets (top-10 units with
three media; top-5 units with one media item per modality) against a single
frozen top-50 manifest, so that retrieval quality is constant by construction
and only the packer's allowance varies. The within-item modality interventions
(\S\ref{sec:protocol:modality}) supply twelve further conditions: four image
variants, four video variants, and four audio variants.

\subsection{LLM-as-Judge Protocol}

All system outputs are evaluated by a single LLM judge (Qwen3.7-Plus; full
prompt in Appendix~\ref{app:prompts}). A three-judge panel (GLM-5.2 and
Grok-4.5 as auxiliaries) re-scores every Track~B run to verify that conclusions
are robust to judge identity (agreement statistics:
Appendix~\ref{app:cross-judge}). Only the primary judge's scores appear in any
reported number. The judge receives the question, expected answer, accepted
aliases, and the system's predicted answer, and assigns a score on a 0--1
scale:

\begin{table*}[!ht]
\centering
\small
\setlength{\tabcolsep}{3pt}
\begin{tabular}{@{}clp{0.55\textwidth}@{}}
\toprule
\textbf{Score} & \textbf{Label} & \textbf{Definition} \\
\midrule
1.0 & Perfect & Semantically identical to expected or alias. Only formatting/capitalization differs. \\
0.7 & Mostly Correct & Right answer but extra context or slightly imprecise wording. \\
0.5 & Partial & Gets category or one part right, misses specifics. \\
0.3 & Attempted & Core fact wrong but shows effort. \\
0.0 & Wrong & Completely incorrect, refused on answerable question, empty, or error. \\
\bottomrule
\end{tabular}
\caption{LLM-as-Judge scoring rubric.}
\label{tab:judge-rubric}
\end{table*}

For \textsc{Answer Refusal} items, ``I don't know'' or similar refusals score 1.0; any other answer scores 0.0.

\subsection{Metrics Definitions}

\begin{itemize}[leftmargin=*,itemsep=2pt]
\item \textbf{Judge score (answerable):} Mean judge score over the 784 answerable items. Answer Refusal is excluded and reported separately as abstention.
\item \textbf{Abstention (paired):} Correct-refusal rate on the 139 \textsc{Answer Refusal} items together with the false-refusal rate on the 784 answerable items (Table~\ref{tab:endtoend}, Figure~\ref{fig:calibration}); reporting both prevents an always-abstain system from appearing well-calibrated.
\item \textbf{Required-fact Recall@$k$:} Fraction of an item's gold required facts whose source unit appears in the top-$k$ set, computed over both the ranked units and the units actually serialized into the request.
\item \textbf{Complete-chain Recall@$k$ / chain delivery:} Fraction of \emph{items} for which \emph{every} required fact appears, ranked or delivered respectively. This is the retrieval metric that matters for multi-hop memory, since a partial chain is not usable.
\item \textbf{Paired effect:} For a controlled contrast, the mean within-item difference over the shared QID set, with an item-bootstrap 95\% interval (10{,}000 resamples), wins/ties/losses, and a persona-clustered bootstrap as a dependence sensitivity check.
\end{itemize}

\section{Dataset Statistics}
\label{app:dataset-stats}

The final MOSAIC-Eval set contains 923 items across 70 personas, averaging 13.2 items per persona and 17.5 sessions per item (range 11--24). The total haystack comprises approximately 18.8M tokens, with a per-persona average of $\sim$269K tokens. Multimodal artefact counts by evidence tier are in Table~\ref{tab:dataset-stats}.

\begin{table}[!ht]
\centering
\small
\setlength{\tabcolsep}{4pt}
\begin{tabular}{@{}lrrrr@{}}
\toprule
\textbf{Tier} & \textbf{Image} & \textbf{Video} & \textbf{Audio} & \textbf{Total} \\
\midrule
Required           &   911 & 195 &   476 &  1{,}582 \\
Evidence (non-req) &   861 & 366 & 1{,}127 &  2{,}354 \\
Filler             & 2{,}525 &   0 & 4{,}294 &  6{,}819 \\
\midrule
Total              & 4{,}297 & 561 & 5{,}907 & 10{,}765 \\
\bottomrule
\end{tabular}
\caption{Multimodal artefacts in MOSAIC.}
\label{tab:dataset-stats}
\end{table}

\begin{table}[!ht]
\centering
\small
\setlength{\tabcolsep}{4pt}
\begin{tabular}{@{}lr@{}}
\toprule
\textbf{Metric} & \textbf{Value} \\
\midrule
Total items & 923 \\
Personas & 70 \\
Items per persona (mean) & 13.2 \\
Answerable items & 784 \\
\textsc{Answer Refusal} items & 139 \\
Sessions per item (mean) & 17.5 \\
Sessions per item (min--max) & 11--24 \\
\midrule
Evidence tokens (total) & 1,861,163 \\
Filler tokens (total) & 16,963,999 \\
Total haystack tokens & 18,825,162 \\
Per-persona tokens (mean) & 268,931 \\
Evidence tokens per session (mean) & 840 \\
Filler tokens per session (mean) & 1213 \\
Answer words (mean) & 4.9 \\
\bottomrule
\end{tabular}
\caption{MOSAIC dataset statistics.}
\label{tab:dataset-stats-app}
\end{table}

MOSAIC's per-persona haystack ($\sim$269K tokens) is substantially larger than
LongMemEval's ($\sim$158K). The token distribution is deliberately skewed:
evidence-bearing sessions contribute a mean of 840 tokens each, while filler
sessions average 1{,}213 tokens (overall session mean $\sim$1.2K). Required
facts are therefore scattered across multimodal evidence and multi-topic
filler rather than clustered in a handful of salient messages. LongMemEval
spreads its smaller budget across more sessions ($\sim$50 at $\sim$3.2K tokens
each) that typically centre on a single text topic, so the retrieval challenge
differs in both scale and modality mix.

\begin{table}[!ht]
\centering
\small
\setlength{\tabcolsep}{4pt}
\begin{tabular}{@{}lrrr@{}}
\toprule
\textbf{Question Type} & \textbf{Count} & \textbf{\%} \\
\midrule
  Knowledge Update & 193 & 20.9\% \\
  Temporal Reasoning & 189 & 20.5\% \\
  Information Extraction & 178 & 19.3\% \\
  Answer Refusal & 139 & 15.1\% \\
  Msr Bridge & 112 & 12.1\% \\
  Msr Selection & 112 & 12.1\% \\
\midrule
\textbf{Total} & \textbf{923} & \textbf{100\%} \\
\bottomrule
\end{tabular}
\caption{Question type distribution.}
\label{tab:question-type-dist}
\end{table}

The six question types are not uniformly distributed. Knowledge Update (20.9\%) and Temporal Reasoning (20.5\%) are the largest categories, reflecting their importance for evaluating evolving state and causal disruption---two capabilities central to long-horizon memory. Information Extraction (19.3\%), the simplest type, serves as a single-hop baseline. The two multi-hop types, MSR Bridge and MSR Selection, each account for 12.1\%, and together with Knowledge Update and Temporal Reasoning, multi-hop reasoning questions constitute 65.6\% of the benchmark. Answer Refusal items (15.1\%) provide the abstention subset, sufficient for measuring refusal behaviour without dominating the evaluation.

\begin{table}[!ht]
\centering
\small
\setlength{\tabcolsep}{4pt}
\begin{tabular}{@{}lrr@{}}
\toprule
\textbf{Modality} & \textbf{All Facts} & \textbf{Required Facts} \\
\midrule
  Text & 914 & 278 \\
  Image & 1,788 & 911 \\
  Audio & 1,304 & 476 \\
  Video & 567 & 195 \\
\midrule
\textbf{Total} & \textbf{4,573} & \textbf{1,860} \\
\bottomrule
\end{tabular}
\caption{Evidence modality distribution. 3,659/4,573 (80.0\%) of all facts are non-text.}
\label{tab:modality-dist}
\end{table}

The modality distribution reflects MOSAIC's deliberate emphasis on non-textual evidence. Across all facts (required + context + distractor), 80.0\% involve a non-text modality, and among required facts specifically, 85.1\% ($1{,}582/1{,}860$) are non-text. Image evidence dominates, accounting for 39.1\% of all facts and 49.0\% of required facts, followed by audio (28.5\% all, 25.6\% required) and video (12.4\% all, 10.5\% required). The relatively lower share of video reflects the higher generation cost and the targeted use of video for temporal gaps (action ordering, state transition, repetition count), where static alternatives cannot suffice. Audio's substantial presence---over a quarter of required facts---is unique among memory benchmarks and directly enables the prosody-based discrimination that Caption-Text architectures consistently fail on (see \S\ref{sec:results}).

\begin{table}[!ht]
\centering
\small
\setlength{\tabcolsep}{4pt}
\begin{tabular}{@{}lr@{}}
\toprule
\textbf{Gap Strategy} & \textbf{Count} \\
\midrule
  Image --- Background Detail & 328 \\
  Image --- Fine Grained Attribute & 388 \\
  Image --- Spatial Configuration & 194 \\
  Video --- Action Ordering & 67 \\
  Video --- Action Repetition Count & 64 \\
  Video --- State Transition & 64 \\
  Audio --- Prosody/Tone (tagged) & 450 \\
\midrule
  Untagged required non-text & 27 \\
\midrule
  \textbf{Total required non-text} & \textbf{1{,}582} \\
\bottomrule
\end{tabular}
\caption{Gap strategy distribution on required facts. Named challenge tags
cover 1{,}555/1{,}582 required non-text facts; the 27 untagged residuals are
26 required audio facts without a tone field and 1 image fact without a
visual strategy (release audit of
\texttt{stage2\_human\_filtered}).}
\label{tab:gap-strategies}
\end{table}

The gap strategy distribution reveals three design priorities. First,
fine-grained attributes (388 required-fact occurrences over 333 items; e.g.,
brand names, prices, or labels) constitute the largest image gap, reflecting
the prevalence of text-bearing visual evidence in real-world omni-modal
interactions. Background detail (328 facts / 284 items) and spatial
configuration (194 / 185) together ensure that models must attend to the full
visual field rather than fixating on salient foreground objects. Second, video
gaps are deliberately balanced across action ordering (67), repetition count
(64), and state transition (64). Third, tagged audio prosody is the single most
frequent gap strategy: 450 required-fact occurrences across 399 answerable
items (50.9\% of 784). Prosody is an intrinsic property of voice messages that
text transcripts discard, so caption-based pipelines are exposed on a large
fraction of the release.

\begin{table}[H]
\centering
\small
\setlength{\tabcolsep}{3pt}
\resizebox{\columnwidth}{!}{%
\begin{tabular}{@{}lrrrr@{}}
\toprule
\textbf{Benchmark} & \textbf{Items} & \textbf{Tok.} & \textbf{Avg.\ Sess.} & \textbf{Non-text} \\
\midrule
MOSAIC (Ours) & 923 & $\sim$269K/pers. & 17.5/item & 80.0\% \\
LongMemEval & 500 & $\sim$158K/pers. & $\sim$50 & 0\% \\
LoCoMo & 1{,}986 & $\sim$17K/conv. & 27.2 & images \\
\bottomrule
\end{tabular}%
}
\caption{Haystack scale comparison with prior conversational memory benchmarks.
LoCoMo figures follow Maharana et~al.\ (ACL 2024): 1{,}986 QA pairs, mean
16.6K tokens and 27.2 sessions per conversation. MOSAIC session counts are
per item; token budgets for MOSAIC/LongMemEval are per persona.}
\label{tab:haystack-comparison}
\end{table}

MOSAIC occupies a distinct position in the benchmark landscape. At
$\sim$269K tokens per persona it is about $1.7\times$ LongMemEval's
per-persona budget ($\sim$158K) and about $16\times$ LoCoMo's
per-conversation budget ($\sim$17K), with a mean of 17.5 sessions per item.
Individual MOSAIC sessions average $\sim$1.2K tokens (evidence 840, filler
1{,}213)---so the scale advantage is the large multimodal haystack around
sparse required facts, not multi-kilotoken sessions. The 80.0\% non-text
fact rate is substantially higher than prior conversational memory suites:
LoCoMo includes shared images in text dialogues but no video or audio, and
LongMemEval is text-only. No prior benchmark combines all four modalities at
this scale.

\section{Pipeline Filtering and Quality Assurance}
\label{app:filtering}

\begin{table}[H]
\centering
\small
\setlength{\tabcolsep}{3pt}
\resizebox{\columnwidth}{!}{%
\begin{tabular}{@{}lrrr@{}}
\toprule
\textbf{Stage} & \textbf{Input} & \textbf{Output} & \textbf{Net} \\
\midrule
Initial generation & --- & 1,050 & --- \\
(1) Structural filter (9 checks) & 1,050 & 1,019 & $-31$ \\
(2) LLM quality judge (4 checks) & 1,019 & 1,004 & $-15$ \\
(3) Omni probe selection & 1,004 & 885 & $-119$ \\
(4) Second quality intersection & 885 & 879 & $-6$ \\
(5) Human-rescued probe FNs & 879 & \textbf{923} & $+44$ \\
\bottomrule
\end{tabular}%
}
\caption{Sample-count lineage. The omni probe checks 864 answerable candidates;
all 140 answer-refusal candidates bypass that probe and enter the next quality
intersection.}
\label{tab:filtering-summary}
\end{table}

The filtering cascade reveals distinct failure profiles at each step. The structural filter (Step~1) catches 31 items (3.0\%) through deterministic rules---primarily duplicate fact values and answer leaks---representing clear-cut generation errors. The LLM quality judge (Step~2) removes 15 additional items (1.5\%). The 1{,}004-item set then comprises 864 answerable and 140 \textsc{Answer Refusal} candidates. The omni probe accepts 745 answerable candidates and rejects 119; adding the 140 bypassed refusal candidates gives 885. A second four-check quality intersection removes five probe-accepted answerable items and one refusal item, yielding the recorded 879-item automated set (740 answerable + 139 refusal). Finally, human review restores 44 of the probe-rejected answerable items, yielding exactly 923 unique items (784 answerable + 139 refusal). Thus the final dataset is a 923-item subset of the 1{,}004-item LLM-filtered set and a 44-item superset of the automated post-review set.

\subsection{LLM Quality Judge: Per-Check Pass Rates}

The LLM quality judge (Step~2) evaluates each of the 1{,}019
structurally-filtered items on four checks; an item advances only if all four
pass ($1{,}019\!\rightarrow\!1{,}004$; Table~\ref{tab:filtering-summary}).
Rates below are from the stored release validation report
(\texttt{llm\_validation\_report.json}).

\begin{table}[H]
\centering
\small
\setlength{\tabcolsep}{3pt}
\begin{tabular}{@{}lrr@{}}
\toprule
\textbf{Check} & \textbf{Pass} & \textbf{Rate} \\
\midrule
Atomicity (one question, one answer) & 1{,}012/1{,}019 & 99.3\% \\
Grounding (answer in evidence) & 1{,}014/1{,}019 & 99.5\% \\
Multimodal required (needs non-text) & 1{,}015/1{,}019 & 99.6\% \\
Gap load-bearing (gaps gate answer) & 1{,}007/1{,}019 & 98.8\% \\
\midrule
Overall (all 4 checks) & 1{,}004/1{,}019 & 98.5\% \\
\bottomrule
\end{tabular}
\caption{LLM quality judge per-check pass rates on the 1{,}019
structurally-filtered items (conjunctive filter; 15 items fail at least one
check).}
\label{tab:llm-quality-checks}
\end{table}

\subsection{Omni Answerability Probe Breakdown}

The omni probe (MiMo-V2.5) checks whether the 864 answerable items in the 1,004-item set are genuinely answerable when given the required evidence in native modality.

\begin{table}[!ht]
\centering
\small
\setlength{\tabcolsep}{4pt}
\begin{tabular}{@{}lr@{}}
\toprule
\textbf{Metric} & \textbf{Value} \\
\midrule
Answerable items checked & 864 \\
Omni probe YES (answerable) & 745 \\
Omni probe NO (unanswerable) & 119 \\
Raw pass rate & 86.2\% \\
\midrule
\textsc{Answer Refusal} candidates bypassed & 140 \\
Probe-YES retained after quality intersection & 740 \\
\textsc{Answer Refusal} retained after intersection & 139 \\
\textbf{Automated post-review set} & \textbf{879} \\
\midrule
Probe-NO items rescued by humans & 44 \\
Candidate-level corrected rate $(745+44)/864$ & 91.3\% \\
\midrule
Final answerable items $(740+44)$ & 784 \\
Final \textsc{Answer Refusal} items & 139 \\
Answerable retention rate $784/864$ & 90.7\% \\
\textbf{Final MOSAIC-Eval set} & \textbf{923} \\
\bottomrule
\end{tabular}
\caption{Omni-probe and final sample-count reconciliation. The 91.3\% statistic
is computed \emph{before} the second quality intersection and is therefore not
a final-set count; the figure quoted in \S\ref{sec:pipeline:stage2} is the
90.7\% retention rate. Neither is a human answerability ceiling, since
annotators adjudicated only a 100-item sample while the probe adjudicated all
candidates.}
\label{tab:omni-probe}
\end{table}

\section{Statistical Power of Modality Interventions}
\label{app:power}

We report minimum detectable effects (MDEs) for the three video nulls in
Appendix~\ref{app:modality-interventions}. Assumptions: two-sided
$\alpha{=}0.05$, 80\% power
($\mathrm{MDE}=(z_{1-\alpha/2}+z_{1-\beta})\,\widehat{\sigma}_{\Delta}/\sqrt{n}$),
paired judge-score deltas under the primary Gemini reader, and the unique-QID
analysis unit (when a QID contributes multiple intervention cases, we average
their deltas). Intervention-case counts are larger for ordering/repetition
because some QIDs host multiple video facts.

\begin{table}[!ht]
\centering
\small
\setlength{\tabcolsep}{3pt}
\begin{tabular}{@{}lrrr@{}}
\toprule
\textbf{Contrast} & \textbf{$n$} & \textbf{Mean $\Delta$} & \textbf{MDE} \\
\midrule
Ordered $-$ shuffled (ordering) & 64 & $-0.009$ & 8.9 \\
Event $-$ frames (repetition) & 56 & $+0.016$ & 10.8 \\
Event $-$ frames (transition) & 64 & $+0.027$ & 11.2 \\
\bottomrule
\end{tabular}
\caption{Paired video MDEs (in judge-score points) at 80\% power. $n$ counts
unique QIDs, the power-analysis unit used in the main text; intervention-case
$n$ values are 67/64/64.}
\label{tab:video-mde}
\end{table}

An MDE does not mean the experiment rules out all larger effects; it means the
study had adequate power only for effects around that size. The nonsignificant
video contrasts therefore leave smaller temporal effects unresolved. Growing
toward $\ge$150 QIDs per video challenge is future work.

\section{Extended Limitations}
\label{app:extended-limitations}

This appendix expands the caveats summarized in \S\ref{sec:limitations}.

\paragraph{Stage associations versus isolated causality.}
Matched stage drops associate each paired contrast with the intervened stage
under a frozen reader, prompt, packer, and judge. Gold versus retrieved
contexts still differ in distractors and composition, so stage labels are
protocol associations rather than fully isolated causal mechanisms.

\paragraph{Underpowered video nulls.}
Video strata are modest ($n{=}56$--$64$ unique QIDs). At 80\% power the MDEs
are $8.9$--$11.2$ points (Appendix~\ref{app:power}), so ordered-versus-shuffled
and related nulls are \emph{inconclusive}, not proof of lossless temporal
compression.

\paragraph{Gold-delivery audio nulls and lexical leakage.}
Raw audio versus ASR ($+0.8$) and ASR plus a gold prosody tag versus ASR
($+1.5$) are near-null under gold delivery for the primary reader. A
model-independent lexical audit of the textualized audio carriers used as the
ASR condition finds frequent paralinguistic cues (e.g., \emph{oh wow},
\emph{ugh}, filled pauses) in the script text, and other gold hops often
already expose the answer value. These nulls therefore do not establish that
the reader recovers sentiment from plain transcripts or raw audio. A separate
correct-versus-swapped symbolic tag intervention does show selector-level
sensitivity (Appendix~\ref{app:modality-interventions}); human-audited
lexically clean subsets defined without conditioning on model failure remain
future work.

\paragraph{Construction and coverage.}
Automated filters and a human audit support release quality
(Appendix~\ref{app:human-eval}). Modality coverage is uneven by design (we
report per-challenge strata), and text-only backends do not yet ingest native
media indices. Gold construction metadata must not be exposed to systems under
test (Appendix~\ref{app:refusal-structure}).

\section{Refusal Surface-Structure Diagnostics}
\label{app:refusal-structure}

\textsc{Answer Refusal} items inherit an IE-shaped template then strip required
evidence. We separate \emph{privileged} construction annotations from
\emph{system-observable} features. Classifiers are balanced logistic regressions
with persona-grouped 5-fold CV; AUROC confidence intervals are item bootstrap
95\% intervals over out-of-fold scores.

\paragraph{Privileged negative control.}
Required-hop count, answer-session count, and the \texttt{answer\_refusal}
question-type label perfectly identify refusal items (AUROC $=1.00$). These
fields are definitionally downstream of refusal construction and are not part
of the evaluation request; this result is a sanity check, \emph{not} an
exploitable surface shortcut. Release users must not feed gold metadata
(\texttt{evidence\_plan} requiredness, \texttt{answer\_session\_ids}, type
labels) into systems under test.

\paragraph{Observable features.}
Haystack length is matched: sessions-only AUROC $0.52$ $[0.47,\,0.56]$ vs.\ all
answerable items. Broader structural counts (messages, media attachments, date
span, question length) reach $0.75$ $[0.70,\,0.79]$. Question-text TF--IDF is
partially predictive ($0.85$ $[0.82,\,0.88]$ vs.\ all answerable; $0.66$
$[0.60,\,0.72]$ vs.\ IE-only). Thus refusal status is not recoverable from
haystack length alone, but residual question-form cues remain; calibration
should not be read as fully form-blind, and matched multi-hop refusal controls
are future work.

\section{Human Evaluation}
\label{app:human-eval}

\subsection{Annotation Protocol}

Three annotators independently reviewed a stratified sample of 100 items (70 where the omni probe verdict was NO, 30 where it was YES). Annotators were shown the question, expected answer, all required evidence facts (with modality and gap strategy), and the omni probe's verdict. For each item, they answered:

\begin{enumerate}[leftmargin=*,itemsep=1pt]
\item \textbf{Well-posed?} Is the question unambiguous and answerable as stated? (Yes / Partially / No)
\item \textbf{Gold answer correct?} Is the expected answer factually correct given the evidence? (Yes / No)
\item \textbf{Evidence sufficient?} Are all required artefacts legible/audible and do they show the answer? (Yes / No)
\item \textbf{Failure mode:} If insufficient, what is the primary failure? (Evidence item invalid / Model refused incorrectly / Model attended to wrong modality / Model overlooked gap-bearing detail / Judge format mismatch / No need to answer)
\end{enumerate}

A follow-up audit then graded 100 additional probe-YES items drawn from the 740 automated answerable releases (seed 20260801; excluding the 30 YES items already in the first sample), using the same Q1--Q3 rubric with two annotators.

\subsection{Release Label Quality (Primary)}

The first audit \emph{oversamples} probe-NO items (70/100, against a base rate of 119/864 probe-adjudicated answerable candidates), so the pooled rates in Table~\ref{tab:human-eval-pooled} (Q2 69.0\%, Q3 65.3\%) describe that stratified sample---not the released set. Table~\ref{tab:human-eval-stratum} reports rates by probe stratum; Table~\ref{tab:human-eval-release} reweights to the 784 answerable release items ($740$ probe-YES $+\,44$ human-rescued).

\begin{table}[!ht]
\centering
\small
\setlength{\tabcolsep}{3pt}
\begin{tabular}{@{}lrrr@{}}
\toprule
\textbf{Stratum / sample} & \textbf{Q1} & \textbf{Q2} & \textbf{Q3} \\
\midrule
Audit probe-YES ($n{=}30$, 3 ann.) & 97.8\% & 100\% & 94.4\% \\
Audit probe-NO ($n{=}70$, 3 ann.) & 92.9\% & 55.7\% & 52.9\% \\
Follow-up YES ($n{=}100$, 2 ann.) & 98.5\% & 97.0\% & 97.0\% \\
Combined probe-YES ($n{=}130$) & 98.3\% & 97.7\% & 96.4\% \\
\bottomrule
\end{tabular}
\caption{Annotator-mean Yes rates by probe stratum (Q1 well-posed, Q2 gold correct, Q3 evidence sufficient). Combined YES merges the audit YES subset with the disjoint probe-YES follow-up (item score $=$ mean of annotator Yes/No).}
\label{tab:human-eval-stratum}
\end{table}

\begin{table}[!ht]
\centering
\small
\setlength{\tabcolsep}{3pt}
\begin{tabular}{@{}lrr@{}}
\toprule
\textbf{Estimand ($N{=}784$ answerable)} & \textbf{Est.} & \textbf{95\% CI} \\
\midrule
Q1 well-posed & 98.4\% & $[97.3,\,99.5]$ \\
Q2 gold answer correct & 97.8\% & $[96.3,\,99.4]$ \\
Q3 evidence sufficient & 96.6\% & $[95.1,\,98.2]$ \\
\bottomrule
\end{tabular}
\caption{Stratified estimates on the 784 answerable MOSAIC-Eval items. Probe-YES stratum ($N{=}740$) uses the combined $n{=}130$ sample; the 44 rescued items enter at rate 1.0 (human-verified at rescue). CIs use item-score variance with finite-population correction on the YES stratum. The 139 refusal items are excluded (different label).}
\label{tab:human-eval-release}
\end{table}

On the released answerable set, gold-answer correctness is therefore $\approx$97.8\% and evidence sufficiency $\approx$96.6\%---not the unweighted 69\%/65\% of the NO-heavy audit sample. Residual errors in the YES follow-up are concentrated in fine-grained visual counts, illegible price tags, and occasional audio--image cue conflicts (annotator notes), consistent with generative-media ambiguity rather than systematic gold-label drift. Pairwise agreement on the follow-up is 97\% (Q1) and 94\% (Q2/Q3).

\subsection{Stratified Audit Sample (Secondary)}

Tables~\ref{tab:human-eval-pooled}--\ref{tab:human-eval-per-eval} retain the original pooled and per-evaluator counts for the 100-item stratified audit. Per-evaluator variance on Q3 (E2: 52\% vs.\ E1: 68\%, E3: 76\%) reflects genuine ambiguity in judging synthesised multimodal artefacts on the probe-NO-heavy sample.

\begin{table}[!ht]
\centering
\small
\setlength{\tabcolsep}{4pt}
\begin{tabular}{@{}lr@{}}
\toprule
\textbf{Question} & \textbf{Pooled \% (3 evaluators)} \\
\midrule
Q1. Well-posed? --- Yes & 94.3\% \\
Q1. Well-posed? --- Partially & 3.0\% \\
Q1. Well-posed? --- No & 2.7\% \\
\midrule
Q2. Gold answer correct? --- Yes & 69.0\% \\
Q2. Gold answer correct? --- No & 31.0\% \\
\midrule
Q3. Evidence sufficient? --- Yes & 65.3\% \\
Q3. Evidence sufficient? --- No & 34.7\% \\
\bottomrule
\end{tabular}
\caption{Pooled responses on the stratified 100-item audit (70 probe-NO / 30 probe-YES). Not a release estimate; see Tables~\ref{tab:human-eval-stratum}--\ref{tab:human-eval-release}.}
\label{tab:human-eval-pooled}
\end{table}

\begin{table}[!ht]
\centering
\small
\setlength{\tabcolsep}{3pt}
\begin{tabular}{@{}lrrr@{}}
\toprule
\textbf{Question} & \textbf{E1} & \textbf{E2} & \textbf{E3} \\
\midrule
Q1. Well-posed? --- Yes & 94 & 99 & 90 \\
Q1. Well-posed? --- Partially & 4 & 0 & 5 \\
Q1. Well-posed? --- No & 2 & 1 & 5 \\
\midrule
Q2. Gold correct? --- Yes & 65 & 66 & 76 \\
Q2. Gold correct? --- No & 35 & 34 & 24 \\
\midrule
Q3. Evidence sufficient? --- Yes & 68 & 52 & 76 \\
Q3. Evidence sufficient? --- No & 32 & 48 & 24 \\
\bottomrule
\end{tabular}
\caption{Per-evaluator responses on the stratified 100-item audit.}
\label{tab:human-eval-per-eval}
\end{table}

\subsection{Failure Mode Distribution}

Table~\ref{tab:failure-modes-human} breaks down failure modes for items where the omni probe verdict was NO. Evidence item invalid (60.0\%) dominates, reflecting generative media artefacts rather than benchmark design flaws.

\begin{table}[H]
\centering
\small
\setlength{\tabcolsep}{4pt}
\begin{tabular}{@{}lrr@{}}
\toprule
\textbf{Failure Mode (Q4)} & \textbf{Count} & \textbf{\%} \\
\midrule
Evidence item invalid & 126 & 60.0\% \\
Model overlooked gap-bearing detail & 34 & 16.2\% \\
Model refused incorrectly & 31 & 14.8\% \\
Model attended to wrong modality & 16 & 7.6\% \\
Judge format mismatch & 3 & 1.4\% \\
\bottomrule
\end{tabular}
\caption{Failure mode distribution on Judge=NO items (pooled across 3 evaluators).}
\label{tab:failure-modes-human}
\end{table}

The failure mode distribution clarifies \emph{why} the omni probe fails. The dominant category---evidence item invalid (60.0\%)---indicates that the primary bottleneck is media synthesis quality: generated images sometimes render text illegibly, video frames occasionally miscount actions, and audio clips may garble prosodic cues. This is a limitation of current generative models rather than a benchmark design flaw, and as synthesis quality improves, these failures will naturally decrease. The second-largest category, model overlooked gap-bearing detail (16.2\%), reveals that even a capable omni model sometimes attends to the correct modality but misses the specific detail that the gap strategy targets---for example, reading the foreground text in an image while missing the small background label that contains the answer. This is precisely the failure mode that MOSAIC is designed to expose in evaluated systems. Model refused incorrectly (14.8\%) indicates cases where the omni probe said ``I don't know'' despite the evidence being present and legible, likely due to ambiguous question phrasing or insufficient reasoning depth. The remaining categories---wrong modality (7.6\%) and format mismatch (1.4\%)---are minor, confirming that the probe generally attends to the correct evidence source.

\subsection{Probe Rejections Overturned (Rescued Items)}

The human audit identified 44 items where the omni probe verdict was NO but annotators restored them to the final set. The selection rule is: an item is \emph{excluded} only if all three annotators mark the evidence item invalid (26 items); otherwise it is \emph{rescued} (44 items). The 44 rescued items are not unanimous first-pass agreement---first-pass Q2 Yes counts were 35/36/46 across the three annotators across the 70 probe-NO items audited---so ``quality 1.0'' means inclusion under the three-annotator unanimous-exclusion rule, not unanimous first-pass acceptance. Rescued items span all five answerable question types: Information Extraction (5), Knowledge Update (11), MSR Bridge (9), MSR Selection (9), and Temporal Reasoning (10).

\subsection{Inter-Rater Agreement}

Pairwise agreement between evaluators measures the reliability of human judgments:
\textbf{Q1 (Well-posed?):} mean pairwise 94.0\% (E1-E2=95\%, E2-E3=91\%, E1-E3=96\%).
\textbf{Q2 (Gold answer correct?):} mean pairwise 92.7\% (E1-E2=99\%, E2-E3=90\%, E1-E3=89\%).
\textbf{Q3 (Evidence sufficient?):} mean pairwise 84.0\% (E1-E2=84\%, E2-E3=76\%, E1-E3=92\%).

Q3 shows the highest variance (76--92\%), reflecting genuine perceptual ambiguity in judging whether a synthesised image or audio clip ``clearly shows'' the intended detail---a finding consistent with prior work on multimodal quality assessment. Q1 has the highest agreement (94.0\%), confirming that MOSAIC questions are unambiguously well-posed.

\subsection{Probe Rejections Upheld}

The audit confirmed 26 items where the omni probe verdict was NO and evaluators agreed the item was genuinely invalid. These items remain excluded from MOSAIC-Eval.

\section{Judge Validation}
\label{app:cross-judge}

Every number in this paper is produced by a \emph{single} judge
(Qwen3.7-Plus), so that all comparisons are scored by the same model;
comparing scores across different judges is unsound. To justify relying on one
judge we run a \textbf{three-judge panel}: GLM-5.2 and Grok-4.5, both with
extended thinking disabled at temperature 0, independently re-score every item
of all twelve Track~B runs under the identical rubric. This yields
$12 \times 923 = 11{,}076$ triple-scored items. Neither auxiliary judge ever
contributes to a reported score; the panel exists only to test whether the
conclusions survive a change of judge.

Table~\ref{tab:cross-judge-app} reports the pooled result. Agreement is very
high on every pair: quadratic-weighted $\kappa$ never falls below 0.96 on
average, Spearman $\rho$ of the item-level scores never below 0.95, and all
three judges land in the same rubric band on 84.0\% of items. Per-run
variation is modest and concentrated in the two lowest-scoring MemoryOS runs,
where most predictions are refusals and the band distribution is degenerate;
even there the minimum weighted $\kappa$ is 0.884. The most contested question
type is knowledge update ($\kappa = 0.926$), which is expected: it is the type
whose gold answer depends on resolving a later correction, so borderline
partial credit is genuinely ambiguous. Answer refusal is nearly perfectly
agreed ($\kappa = 0.983$) because it reduces to a binary rule.

\paragraph{Human spot-check of LLM judge scores.}
To further validate the primary judge, two human annotators independently scored a stratified 50-item sample of ReAct Native-MM predictions (10 items per rubric band, same scoring rubric). Human--human quadratic-weighted $\kappa = 0.70$ (exact agreement 60\%, within-one-band 88\%). Agreement of each annotator with the primary LLM judge was $\kappa = 0.79$ and $\kappa = 0.69$ respectively (within-one-band 88\% and 90\%; human-mean $\kappa = 0.82$). Disagreements are concentrated in adjacent bands where partial credit is genuinely ambiguous, consistent with the three-LLM-judge pattern above. We report mean partial-credit judge scores throughout rather than exact match.

\begin{table}[H]
\centering
\scriptsize
\setlength{\tabcolsep}{2pt}
\begin{tabular}{@{}lrrr@{}}
\toprule
\textbf{Pair} & \textbf{$\kappa$} & \textbf{Band} & \textbf{$\rho$} \\
\midrule
Qwen vs.\ GLM-5.2  & 0.962 & 0.880 & 0.954 \\
Qwen vs.\ Grok-4.5 & 0.966 & 0.898 & 0.957 \\
GLM-5.2 vs.\ Grok-4.5 & 0.971 & 0.892 & 0.963 \\
\midrule
\multicolumn{4}{@{}l}{\textit{Mean score (12 runs)}} \\
Qwen (primary) & \multicolumn{3}{r}{0.536} \\
GLM-5.2 & \multicolumn{3}{r}{0.544} \\
Grok-4.5 & \multicolumn{3}{r}{0.534} \\
\midrule
Same band (all three) & \multicolumn{3}{r}{0.840} \\
Items pooled & \multicolumn{3}{r}{11{,}076} \\
\midrule
\multicolumn{4}{@{}l}{\textit{Per-type $\kappa$}} \\
IE / KU / TR & \multicolumn{3}{r}{0.958 / 0.926 / 0.966} \\
MSR-B / MSR-S / AR & \multicolumn{3}{r}{0.974 / 0.968 / 0.983} \\
\bottomrule
\end{tabular}
\caption{Three-judge panel agreement, pooled over all twelve Track~B runs
(quadratic-weighted $\kappa$ over the five rubric bands). The three judges
differ by at most 0.010 in mean score and agree on the induced system ranking,
which justifies reporting every result under the single primary judge.}
\label{tab:cross-judge-app}
\end{table}

\section{Additional Results}
\label{app:additional-results}

This appendix holds the numeric modality-intervention table deferred from
\S\ref{sec:results:modality}, paired W/T/L companions to the main-paper
floats, plus per-question-type ladders, retrieval/delivery curves, drop-reason
accounting, per-challenge and per-family breakdowns, and the full Track~B score
table. Every number comes from the same canonical paired run as the main paper.

\subsection{Matched Stage Ladder: Paired Contrasts}
\label{app:stage-ladder}

Numeric companion to Table~\ref{tab:stage-results} and
Figure~\ref{fig:stage-ladder}. Gold symbolic facts score 0.846, bounding
residual integration and grading error at 15.4 points. The
perception-and-integration cost of native evidence (symbolic $-$ native) is
6.2 points [95\% CI 3.2, 9.3; W/T/L 206/489/89]. Generic captions and ASR lose
a further 9.3 points relative to native evidence, of which only 1.4 are
recovered by the query-blind structured writer ($+1.4$ [$-0.9$, 3.7];
103/585/96). Retrieval loss (gold native $-$ retrieved all-delivered) is
7.6 points [5.2, 9.9], and production delivery caps remove another 3.8 on the
identical ranked hits [1.5, 6.0]. The full production pipeline reaches 0.653,
only 1.7 points below the original-query budgeted condition [$-0.6$, 3.9]; gold
symbolic facts exceed it by 19.2 points [15.7, 22.8].

Video-dependent items score 0.900 symbolically but 0.729 with native video
(17.1-point perception cost), against 6.2 for image and none for audio
(0.821$\to$0.826). Probe-pass ($n{=}740$) vs.\ human-rescued ($n{=}44$)
items are indistinguishable under gold symbolic facts (0.846 vs.\ 0.845) but
differ by 27.8 points under gold native evidence (0.799 vs.\ 0.520), persisting
through the production pipeline (15.5 [2.9, 28.5]).

\subsection{Second-Reader Stage Replication}
\label{app:second-reader}

Gemini~3.5 (primary) and MiMo-V2.5 (second reader; also the answerability
probe) on the fixed 120-item pilot, identical evidence packs and judge.
MiMo reproduces the gold and production-budgeted ladder; its uncapped
all-delivered row is dominated by API timeouts ($\sim$50 base64 media units)
and is not used for stage attribution.

\begin{table}[H]
\centering
\scriptsize
\setlength{\tabcolsep}{3pt}
\renewcommand{\arraystretch}{0.9}
\begin{tabular}{@{}lrr@{}}
\toprule
\textbf{Condition} & \textbf{Gemini} & \textbf{MiMo} \\
\midrule
Gold symbolic facts & 0.878 & 0.870 \\
Gold native evidence & 0.822 & 0.818 \\
Gold generic text & 0.709 & 0.739 \\
Gold structured text & 0.738 & 0.707 \\
Retrieved, production-budgeted & 0.694 & 0.669 \\
Retrieved, all delivered$^\dagger$ & 0.713 & 0.157 \\
\bottomrule
\end{tabular}
\caption{Mean judge score on the 120-item pilot ($n{=}120$ per cell).
$^\dagger$MiMo all-delivered is provider-limited (91/120 timeouts) and is
reported only to document the failure mode.}
\label{tab:second-reader}
\end{table}

\subsection{Delivery Budget Sweep}
\label{app:budget}

Companion to Figure~\ref{fig:budget}. Required-fact Recall@50 is 0.871
throughout. Halving the unit budget costs 5.7 points [3.3, 8.0];
one-media-per-modality packing costs 17.7 [14.5, 20.9] and collapses
complete-chain delivery from 0.675 to 0.184 (required-fact coverage
0.871$\to$0.501).

\subsection{Within-Item Modality Intervention Contrasts}
\label{app:modality-interventions}

Numeric companion to Figure~\ref{fig:modality} in \S\ref{sec:results:modality}.

\begin{table}[H]
\centering
\scriptsize
\setlength{\tabcolsep}{2pt}
\renewcommand{\arraystretch}{0.9}
\begin{tabular}{@{}llrr@{}}
\toprule
\textbf{Mod.} & \textbf{Contrast} & \textbf{$N$} &
\textbf{$\Delta$ [95\% CI]} \\
\midrule
Img & Raw $-$ OCR (fine/bg) & 716 & $+0.185$ [0.153, 0.216] \\
Img & Raw $-$ OCR (spatial) & 194 & $+0.320$ [0.248, 0.392] \\
Img & Spatial $-$ object-only & 194 & $+0.154$ [0.095, 0.215] \\
Vid & Ordered $-$ shuffled & 67 & $-0.015$ [$-0.078$, 0.046] \\
Vid & Event $-$ frames (rep.) & 64 & $+0.014$ [$-0.052$, 0.080] \\
Vid & Event $-$ frames (trans.) & 64 & $+0.027$ [$-0.050$, 0.106] \\
Aud & Raw $-$ ASR & 450 & $+0.008$ [$-0.015$, 0.030] \\
Aud & Gold prosody $-$ ASR & 450 & $+0.015$ [$-0.008$, 0.038] \\
Aud & Correct prosody $-$ swapped & 399 & $+0.046$ [0.017, 0.077] \\
\bottomrule
\end{tabular}
\caption{Within-item modality interventions (governing stratum).
$N$ is intervention cases except correct$-$swapped, which uses unique QIDs
(primary analysis unit). W/T/L: fine/bg 222/434/60; spatial raw$-$OCR 92/89/13;
spatial desc.$-$object 58/117/19; ordered$-$shuffled 5/56/6; repetition 6/54/4;
transition 11/44/9; raw$-$ASR 36/387/27; prosody-tag 34/394/22;
correct$-$swapped 46/330/23.}
\label{tab:modality-interventions}
\end{table}

\paragraph{Prosody-tag selector sensitivity.}
We reran both arms under one frozen reader, prompt, gold-delivery packer, and
judge: \emph{ASR + correct gold tone} versus the identical context with
\texttt{happy}$\leftrightarrow$\texttt{unhappy} flipped. The eligible set is
all prosody carriers (450 intervention cases spanning 399 unique QIDs; 51 QIDs
host two carriers). We do \emph{not} condition on ASR failure. The primary
unit is the unique-QID mean paired delta: $+0.046$ $[0.017,\,0.077]$
(W/T/L 46/330/23; persona-clustered mean $+0.041$ over 70 personas). Condition
means are 0.824 (correct) and 0.782 (swapped). The case-level mean delta is
$+0.043$---slightly below the QID mean because multi-carrier QIDs are
down-weighted when averaging by QID. A positive correct-over-swapped effect
shows that the \emph{annotated} prosodic selector is causally load-bearing; it
does not show that raw audio perception outperforms ASR.

\subsection{Track~A Fixed-Reader Paired Contrasts}
\label{app:track-a}

Companion to Table~\ref{tab:track-a}. A-MEM beats same-representation flat RAG
by 16.2 points on caption text [13.1, 19.3; W/T/L 233/464/87] and by 16.4 on
omni text [13.2, 19.4]. MemoryOS is worse than flat RAG by 25.5 / 22.1 points
on caption / omni text (315 / 300 paired losses). A-MEM chain coverage is 0.787
vs.\ 0.698 (flat caption) and 0.466 (native multimodal).

\subsection{Track~B End-to-End Scores}
\label{app:endtoend}

Full numeric companion to Figure~\ref{fig:calibration} in
\S\ref{sec:results:endtoend}.
Table~\ref{tab:endtoend} lists every Track~B configuration; per-system and
per-type plots follow in Appendix~\ref{app:visualizations}.

\begin{table}[H]
\centering
\tiny
\setlength{\tabcolsep}{1.5pt}
\renewcommand{\arraystretch}{0.88}
\begin{tabular}{@{}llrrrrr@{}}
\toprule
\textbf{Back-end} & \textbf{Repr.} & \textbf{Ans.} &
\textbf{C.ref} & \textbf{F.ref} &
\textbf{Lat.} & \textbf{Rdr} \\
\midrule
Naive RAG & Caption & 0.390 & 0.835 & 0.431 & 5.8 & DS \\
Naive RAG & Omni & 0.375 & 0.871 & 0.421 & 6.2 & DS \\
Naive RAG & Native-MM & 0.614 & 0.734 & 0.143 & 29.6 & MiMo \\
Naive RAG & Native-MM & 0.687 & 0.496 & \textbf{0.022} & 31.2 & Gem. \\
\midrule
ReAct & Caption & 0.415 & 0.820 & 0.404 & 18.6 & DS \\
ReAct & Omni & 0.419 & 0.799 & 0.390 & 17.5 & DS \\
ReAct & Native-MM & \textbf{0.732} & 0.410 & \textbf{0.018} & 73.6 & MiMo \\
ReAct & Native-MM & 0.674 & 0.417 & 0.037 & 46.4 & Gem. \\
\midrule
A-MEM & Caption & 0.623 & 0.719 & 0.173 & -- & DS \\
A-MEM & Omni & 0.611 & 0.763 & 0.177 & -- & DS \\
MemoryOS & Caption & 0.243 & \textbf{0.885} & 0.649 & 205.4 & DS \\
MemoryOS & Omni & 0.255 & \textbf{0.892} & 0.614 & 228.4 & DS \\
\bottomrule
\end{tabular}
\caption{Track~B on 923 items (vs.\ Figure~\ref{fig:calibration}).
Ans./C.ref/F.ref/Lat.~=~answerable, correct/false refusal, latency (s).
Rdr: DS=DeepSeek-v4-Flash, MiMo=MiMo-V2.5, Gem.=Gemini~3.5.}
\label{tab:endtoend}
\end{table}

\subsection{Per-Question-Type Diagnostic Ladder}
\label{app:extended-stage}

Table~\ref{tab:extended-stage} decomposes the matched ladder of
Table~\ref{tab:stage-results} by question type. The stage profile is
type-dependent in an interpretable way. Knowledge-update items lose the most
when symbolic facts are replaced by native evidence (0.898 to 0.668), since
resolving which of two conflicting states is current depends on perceiving
both. \textsc{Msr-Bridge} items instead degrade mainly at retrieval and
delivery (0.790 gold native to 0.507 end-to-end), because a bridge fails
whenever \emph{either} hop is missing from the request. Information extraction
is nearly flat across the ladder, and temporal reasoning holds up better than
its reputation would suggest once the required evidence is actually delivered.

\begin{table*}[!t]
\centering
\small
\setlength{\tabcolsep}{5pt}
\begin{tabular}{@{}lrrrrrrc@{}}
\toprule
\textbf{Condition} & \textbf{All} & \textbf{IE} & \textbf{MSR-S} &
\textbf{MSR-B} & \textbf{KU} & \textbf{TR} & \textbf{95\% CI} \\
\midrule
Gold symbolic facts & 0.846 & 0.730 & 0.741 & 0.924 & 0.898 & 0.916 & [0.822, 0.869] \\
Gold native evidence & 0.783 & 0.799 & 0.777 & 0.790 & 0.668 & 0.885 & [0.759, 0.806] \\
Gold generic text & 0.690 & 0.688 & 0.660 & 0.665 & 0.590 & 0.826 & [0.662, 0.718] \\
Gold structured text & 0.704 & 0.688 & 0.675 & 0.737 & 0.611 & 0.811 & [0.677, 0.731] \\
Retrieved native, all & 0.708 & 0.774 & 0.683 & 0.589 & 0.596 & 0.843 & [0.679, 0.734] \\
Retrieved native, capped & 0.670 & 0.734 & 0.712 & 0.562 & 0.569 & 0.752 & [0.642, 0.698] \\
End-to-end native RAG & 0.653 & 0.747 & 0.674 & 0.507 & 0.551 & 0.743 & [0.624, 0.682] \\
\bottomrule
\end{tabular}
\caption{Per-question-type diagnostic ladder on the 784 answerable items
(IE $n{=}178$, MSR-S 112, MSR-B 112, KU 193, TR 189); refusal metrics are
reported separately in Table~\ref{tab:endtoend}.}
\label{tab:extended-stage}
\end{table*}

\subsection{Retrieval and Delivery Curves}
\label{app:recall-curves}

Table~\ref{tab:recall-curves} reports required-fact and complete-chain recall
over the ranked units. Complete-chain recall is far below fact recall at every
cutoff, which is precisely why chain-level accounting rather than fact-level
recall is the correct retrieval metric for multi-hop memory: a chain is only
useful if \emph{all} of its members arrive. Query rewriting costs roughly one
point of recall at every cutoff. After packing, these rankings deliver complete
chains for 0.675 (all feasible hits), 0.491 (production caps), and 0.464
(end-to-end) of items, with required-fact delivery coverage 0.871, 0.761, and
0.742 respectively.

\begin{table}[H]
\centering\small
\setlength{\tabcolsep}{3pt}
\begin{tabular}{@{}lrrrrr@{}}
\toprule
\textbf{Metric} & \textbf{@3} & \textbf{@5} & \textbf{@10} &
\textbf{@20} & \textbf{@50} \\
\midrule
\multicolumn{6}{@{}l}{\emph{Original-question ranking (retrieved conditions)}} \\
Required-fact recall & 0.600 & 0.700 & 0.786 & 0.828 & 0.871 \\
Complete-chain recall & 0.282 & 0.388 & 0.515 & 0.583 & 0.676 \\
\multicolumn{6}{@{}l}{\emph{Production fixed-rewrite ranking (end-to-end)}} \\
Required-fact recall & 0.574 & 0.677 & 0.759 & 0.806 & 0.863 \\
Complete-chain recall & 0.265 & 0.371 & 0.490 & 0.556 & 0.661 \\
\bottomrule
\end{tabular}
\caption{Required-fact and complete-chain recall over ranked units, on the 784
answerable items.}
\label{tab:recall-curves}
\end{table}

\subsection{Why Delivery Fails}
\label{app:drop-reasons}

Table~\ref{tab:drop-reasons} accounts for every artefact that was retrieved but
never serialized into the reader's request, counted from the per-request
delivery manifests of the \emph{retrieved native, production caps} condition
(top-20 units, three media per modality;
Appendix~\ref{app:diagnostic-conditions}). Across the 784 answerable items this
amounts to 9{,}338 dropped artefacts---5{,}559 images, 593 videos, and 3{,}186
audio clips. The result is unusually clean: \emph{every} dropped artefact is
attributable to the per-modality media-count cap. No chain member was lost to
token limits, serialization failures, or unsupported formats. Delivery loss in
MOSAIC is therefore a budgeting decision rather than an infrastructure failure,
which is what makes it directly actionable.

\begin{table}[H]
\centering\small
\setlength{\tabcolsep}{4pt}
\begin{tabular}{@{}lrrrr@{}}
\toprule
\textbf{Drop reason} & \textbf{Image} & \textbf{Video} & \textbf{Audio} &
\textbf{All} \\
\midrule
Media-count cap & 5{,}559 & 593 & 3{,}186 & 9{,}338 \\
Context/token cap & 0 & 0 & 0 & 0 \\
Serialization/API error & 0 & 0 & 0 & 0 \\
Unsupported format & 0 & 0 & 0 & 0 \\
\midrule
Total dropped artefacts & 5{,}559 & 593 & 3{,}186 & 9{,}338 \\
\bottomrule
\end{tabular}
\caption{Retrieved-but-dropped artefacts by modality and recorded drop reason,
under production delivery caps (media-count cap applied per modality). The
end-to-end condition shows the same single-cause pattern (6{,}184 image, 587
video, 2{,}533 audio; 9{,}304 total).}
\label{tab:drop-reasons}
\end{table}

\subsection{Per-Challenge Accuracy}
\label{app:gap-strategy}

Table~\ref{tab:gap-strategy-accuracy} reports the best end-to-end system's
judge score by required challenge. Read alongside the within-item interventions
(\S\ref{sec:results:modality}), this localizes which signals survive in a
deployed system rather than under gold delivery. Fine-grained attributes score
highest (0.801) because the native channel resolves them directly once
delivered, whereas background detail is the weakest image challenge (0.670):
the target is by construction not the salient subject, so it survives neither
captioning nor a reader's attention budget. Repetition count is the weakest
challenge overall (0.666), consistent with the video perception cost measured
in \S\ref{sec:results:stages}. On the prosody stratum ($N{=}399$), the same
per-challenge accounting gives 0.361 for Track~B Caption-Text Naive RAG,
0.715 for Naive RAG Native-MM (Gemini~3.5), and 0.740 for ReAct Native-MM
(MiMo-V2.5)---the contrast quoted in \S\ref{sec:results:modality}.

\begin{table}[H]
\centering\small
\setlength{\tabcolsep}{4pt}
\begin{tabular}{@{}lrr@{}}
\toprule
\textbf{Challenge (required)} & \textbf{Score} & \textbf{N} \\
\midrule
Image --- Background detail       & 0.670 & 284 \\
Image --- Fine-grained attribute  & 0.801 & 333 \\
Image --- Spatial configuration   & 0.714 & 185 \\
Video --- Action ordering         & 0.773 & 64 \\
Video --- Action repetition count & 0.666 & 56 \\
Video --- State transition        & 0.697 & 64 \\
Audio --- Prosody                 & 0.740 & 399 \\
\bottomrule
\end{tabular}
\caption{Per-challenge judge score for the best end-to-end system (ReAct over
the native MiMo-V2.5 stack). An item may contribute to multiple rows, so the
strata overlap and do not sum to 784.}
\label{tab:gap-strategy-accuracy}
\end{table}

\subsection{Method Family Ranking}
\label{app:family-ranking}

Table~\ref{tab:family-ranking} groups the twelve Track~B configurations into
architecture families. Text-memory families (A-MEM, MemoryOS) never ingest
native media and use a different retrieval substrate from the RAG families, so
rankings vs.\ Native-MM reflect substrate as well as architecture. The
fixed-reader track (\S\ref{sec:results:architecture}) removes that confound
and partly inverts the ordering---which is why we report both tracks.

\begin{table}[H]
\centering\small
\setlength{\tabcolsep}{4pt}
\begin{tabular}{@{}lrr@{}}
\toprule
\textbf{Family} & \textbf{Mean score} & \textbf{Range} \\
\midrule
Native-MM ReAct              & 0.703 & 0.674--0.732 \\
Native-MM RAG                & 0.651 & 0.614--0.687 \\
A-MEM (text architectures)   & 0.617 & 0.611--0.623 \\
Text-only ReAct              & 0.417 & 0.415--0.419 \\
Text-only RAG                & 0.383 & 0.375--0.390 \\
MemoryOS (text architectures) & 0.249 & 0.243--0.255 \\
\bottomrule
\end{tabular}
\caption{Method family ranking by mean answerable judge score across the twelve
Track~B runs (two configurations per family).}
\label{tab:family-ranking}
\end{table}

\subsection{Per-System and Per-Question-Type Visualizations}
\label{app:visualizations}
\label{sec:results:per-type}

\begin{figure*}[t]
\centering
\includegraphics[width=0.92\textwidth]{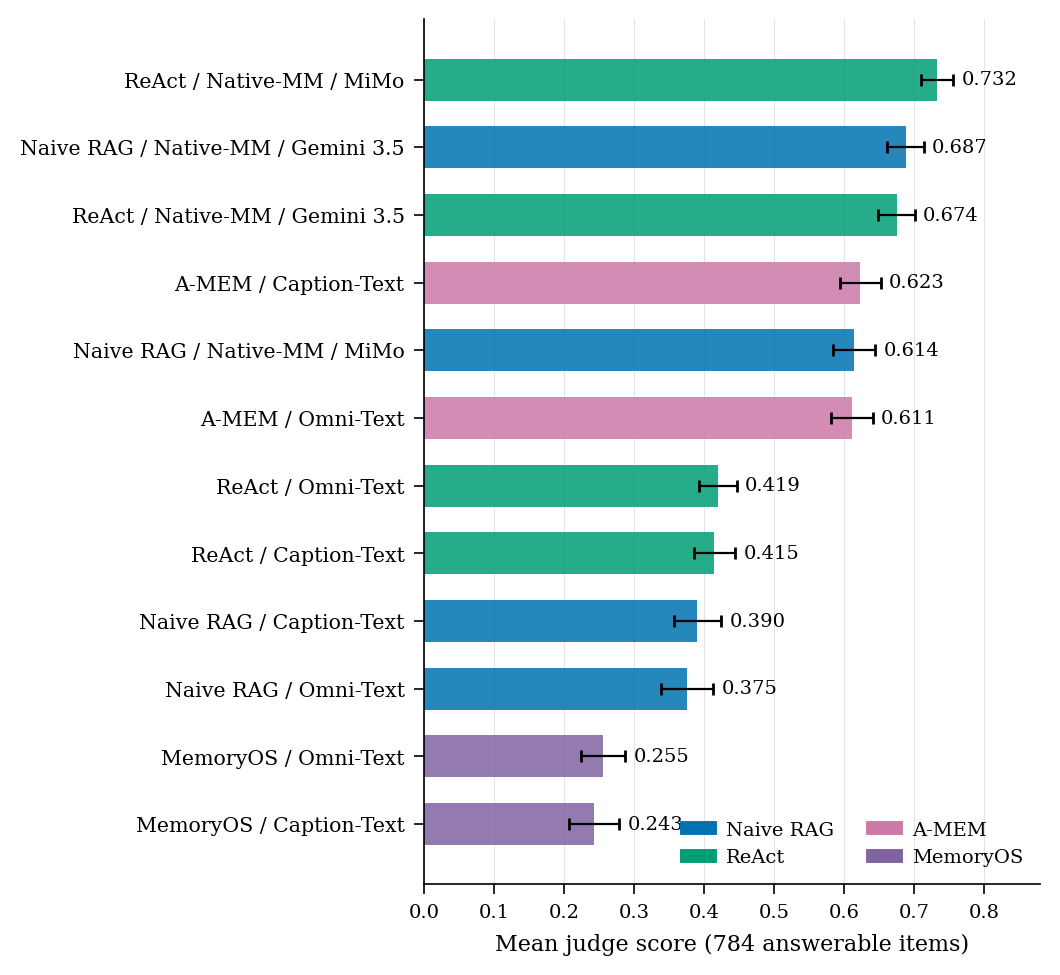}
\caption{Mean judge score on the 784 answerable items for every
Track~B system, with persona-clustered bootstrap 95\% intervals. Colour encodes
the architecture family. The native-multimodal configurations occupy the top of
the ranking, and the two MemoryOS configurations the bottom.}
\label{fig:overall_score}
\end{figure*}

\begin{figure*}[t]
\centering
\includegraphics[width=0.72\textwidth]{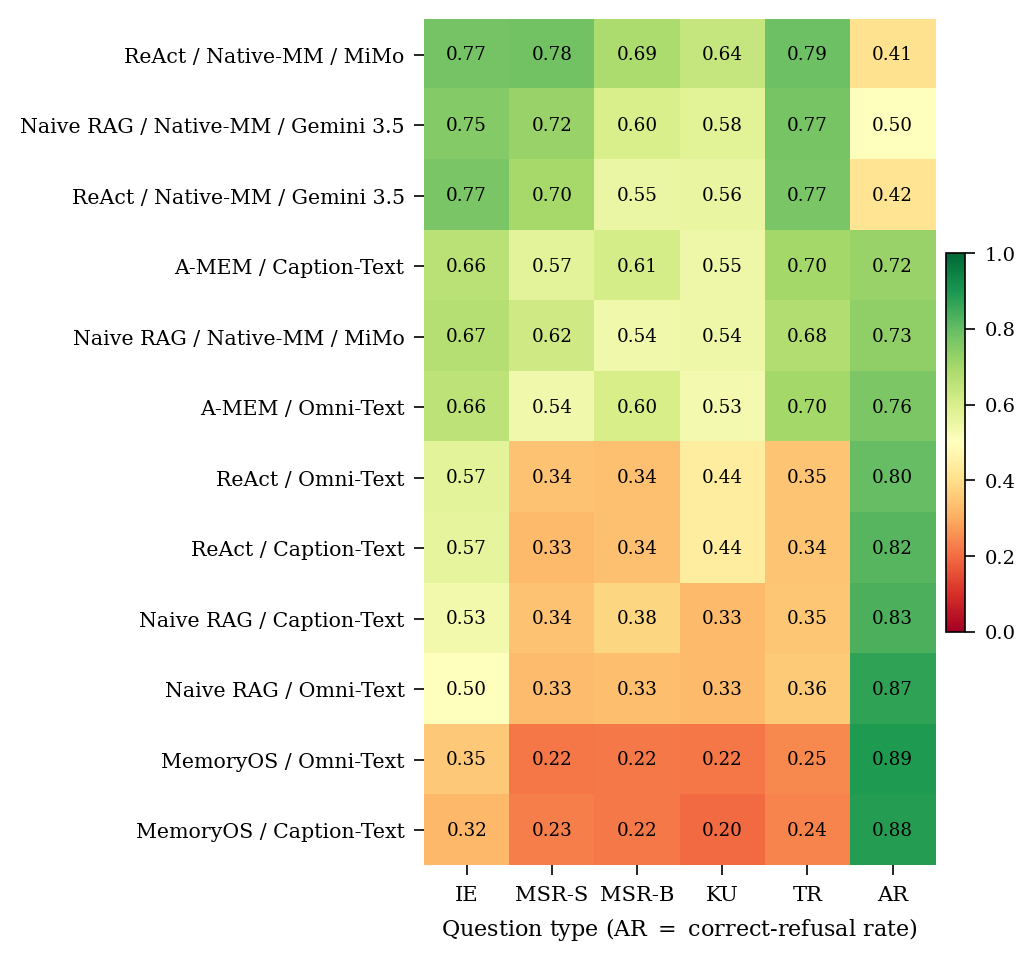}
\caption{Judge score by system $\times$ question type, with
\textbf{AR} the correct-refusal rate on the 139 refusal items. Knowledge Update
is the hardest answerable type for essentially every system. The AR column runs
\emph{opposite} to every answerable column, which is the
indiscriminate-abstention pattern quantified in \S\ref{sec:results:endtoend}.}
\label{fig:heatmap}
\end{figure*}

\begin{figure*}[t]
\centering
\includegraphics[width=\textwidth]{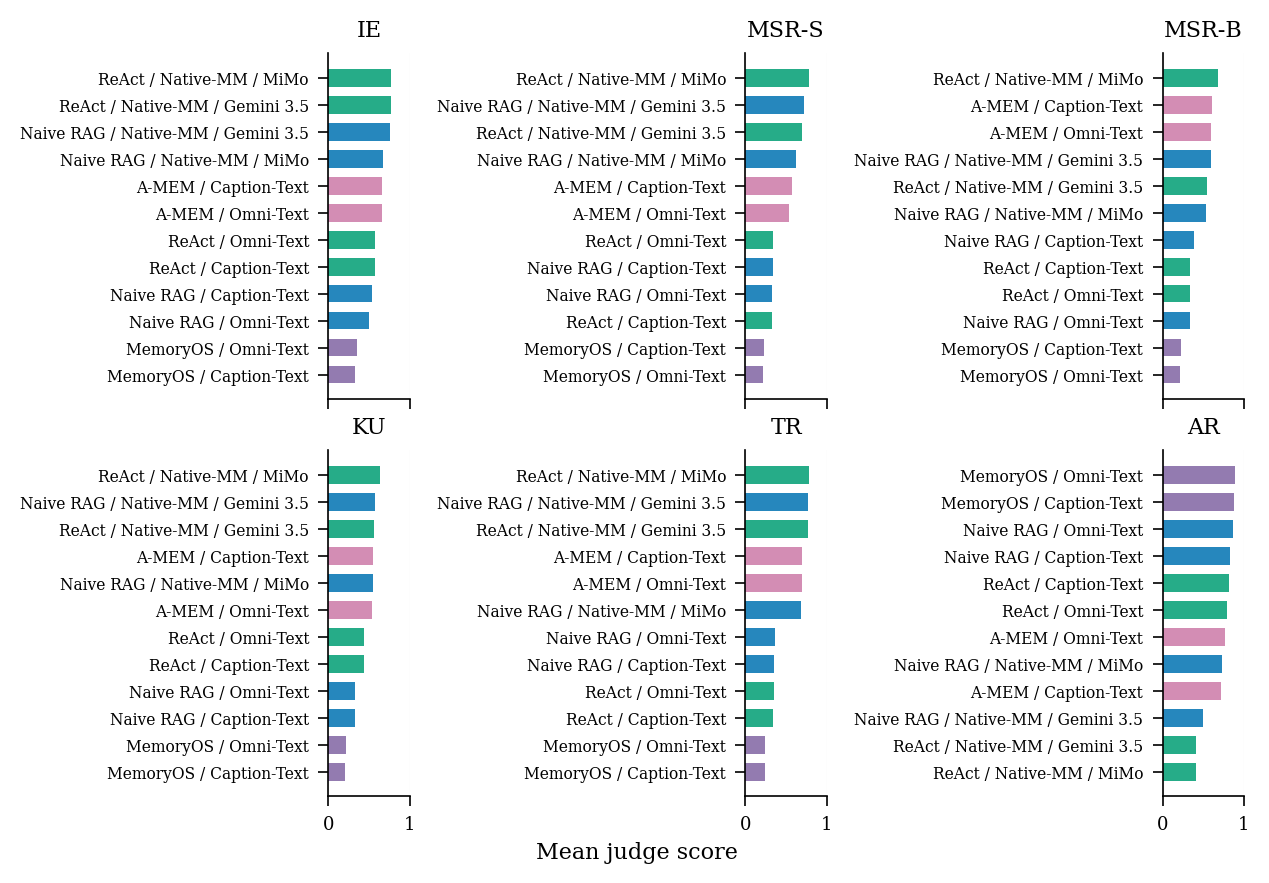}
\caption{Judge score by question type: six panels, each ranking all
twelve systems for that type. Colour encodes the architecture family. The
answerable panels share a near-identical ordering, while the AR panel reverses
it.}
\label{fig:by_question_type}
\end{figure*}

Figures~\ref{fig:overall_score}--\ref{fig:by_question_type} visualize
Table~\ref{tab:endtoend}: overall ranking, system$\times$type heatmap
(including AR), and per-type panels.

\section{Qualitative Examples}
\label{app:qualitative}

\subsection{Example Item: Multi-Hop Cross-Modal Chain (MSR Bridge)}

\textbf{Question:} \textit{``What is the packing method from the carver in the video where I first hollowed then inlaid, in the town on the sign in my chess set photo, that I sounded happy about?''}

\textbf{Expected Answer:} \textit{``bundle-wrap''}

\textbf{Reasoning Chain:}
\begin{enumerate}[leftmargin=*,itemsep=1pt]
\item \textbf{Selector} (Image, Background Detail): The user shares a photo of their driftwood chess set on a hatch-cover table mid-game. A sign visible through the window in the background reads ``Mistport''---this identifies the town where the carver's workshop is located.
\item \textbf{Selector} (Video, Action Ordering): A video from a Mistport workshop shows a carver working on driftwood pieces---first hollowing, then inlaying. The video gap requires perceiving the action sequence (hollow $\rightarrow$ inlay) to identify the correct carver among two at the workshop.
\item \textbf{Answer} (Audio, Happy Tone): A voice message mentions two packing methods (``bundle-wrap'' and ``crate-pack''), with happy tone toward ``bundle-wrap''. The text transcript alone names both methods neutrally; only the prosody discriminates.
\end{enumerate}

\textbf{Modalities required:} Image (chess set with background sign) $\rightarrow$ Video (carving action sequence) $\rightarrow$ Audio (happy tone toward packing method). A text-only system would see captions for all three but lose both the spatial background detail and the prosodic discriminator.

\subsection{Example of Model Success}

\textbf{Question:} \textit{``In the photo of my fondue set with the colour-coded forks, what brand is stamped on the base?''}

\textbf{Expected:} \textit{``Lorraine''}

\textbf{Predicted (RAG MM MiMo):} \textit{``Lorraine''} \hspace{1em} \textbf{Score: 1.0}

The model correctly retrieved the image chunk (a photo of a vintage cast-iron fondue set) and read the brand name ``Lorraine'' stamped on the base. The brand was never mentioned in the chat text---only visible in the image.

\subsection{Case Study: Audio Prosody Gap (MSR Selection)}

Figure~\ref{fig:case-study-binoculars} shows the evidence image from
\texttt{mm\_msr\_selection\_ac58aedd\_002}. The comparison below illustrates a
Track~B failure mode discussed in \S\ref{sec:results}: Caption-Text loses the
labeled prosodic discriminator and selects the wrong entity; Native-MM
preserves it. Transcripts below are the released required-evidence utterances
(not paraphrases); see the selector note on the imperfect lexical cue.

\begin{center}
\includegraphics[width=0.55\linewidth]{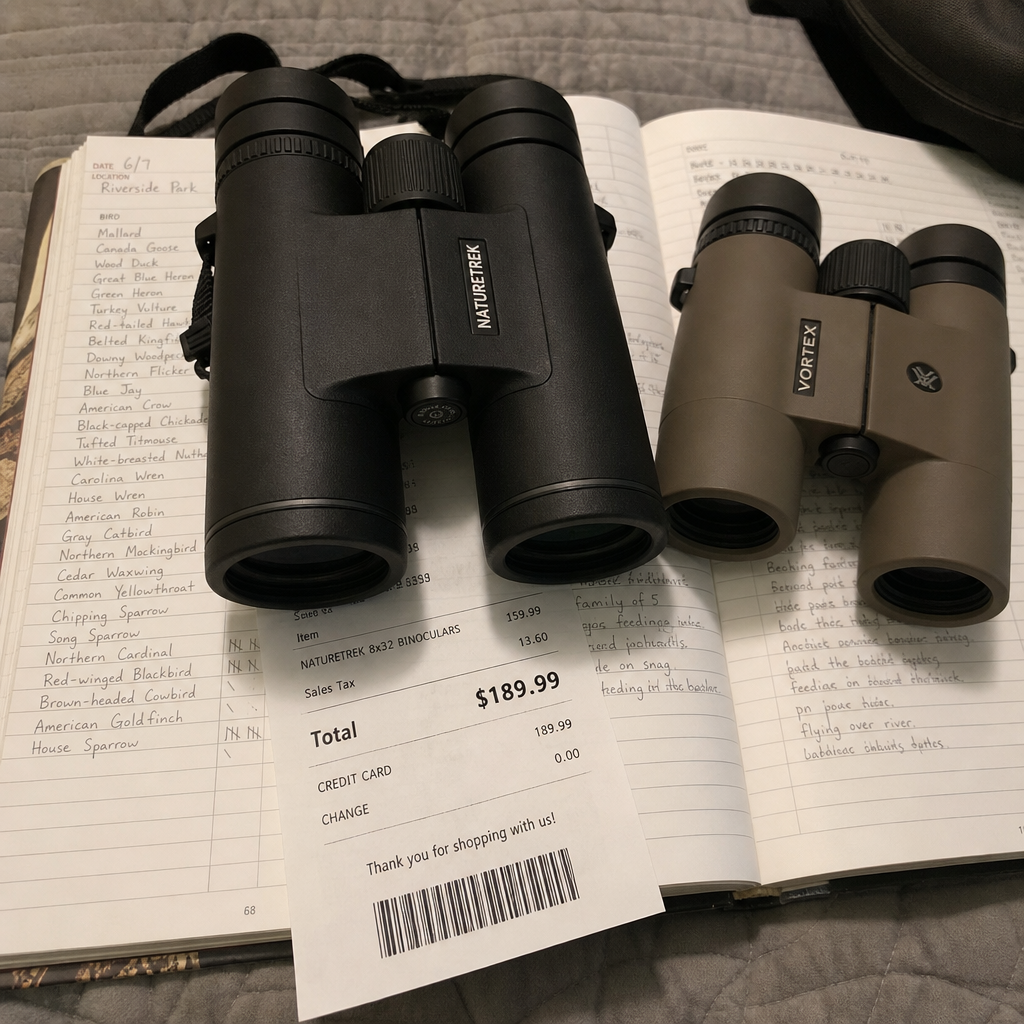}
\captionof{figure}{Birding-log photo from item
\texttt{mm\_msr\_selection\_ac58aedd\_002}. The answer-critical detail
(receipt price \$189.99 under the black 10x42 pair) is only readable in the
image; the chat text never states the price.}
\label{fig:case-study-binoculars}
\end{center}

\textbf{Question:} \textit{``What was the price of the binoculars I sounded happier about?''}

\textbf{Expected Answer:} \textit{``\$189.99''} (black 10x42 pair, identified by happy tone)

\textbf{Chain structure (MSR Selection):}
\begin{enumerate}[leftmargin=*,itemsep=1pt]
\item \textbf{\textsc{Selector}} (Audio, tone contrast): Required voice messages name both candidate pairs without an explicit prefer/like lexicon:
\begin{itemize}[leftmargin=*,itemsep=0pt]
\item ``I also tried the Nikon Monarch 5 8x42 while I was there, and then I gave the mossy oak 8x42 roof prism a go right after.''
\item ``So I went back and tried the other two binoculars I was looking at---the Mossy Oak camo 8x42s and the black 10x42s. The black pair really surprised me.''
\end{itemize}
Construction labels assign unhappy tone to the Mossy Oak 8x42s and happy tone to the black 10x42s. The phrase ``surprised me'' lexically singles out the black pair as the reacted-to entity, but it does not explicitly encode positive polarity (e.g., \emph{prefer}/\emph{happy}); we therefore treat this item as an imperfect Track~B illustration of selector failure under ASR, not as a pure lexical-neutrality exemplar.
\item \textbf{\textsc{Answer}} (Image, fine-grained attribute): A photo of both pairs on a birding log with the receipt tucked under the black pair. The price is never spoken---only readable in the image.
\end{enumerate}

\begin{tcolorbox}[colback=black!5!white,colframe=black!75!black,title={Caption-Text Naive RAG (Whisper ASR + VL Captioner) --- Wrong}]
\textbf{What the system sees:} ASR keeps the wording above (including the entity cue ``surprised me'') but drops delivery/prosody; the VL captioner describes both binoculars and a receipt without binding preference polarity to the correct pair.\\
\textbf{Result:} The selector is misresolved, so the model reads the wrong receipt value. \textbf{Predicted:} \textit{``\$159.99''} (incorrect).
\end{tcolorbox}

\begin{tcolorbox}[colback=blue!5!white,colframe=blue!75!black,title={Native-MM Naive RAG (Gemini~3.5, raw audio + image) --- Correct}]
\textbf{What the system sees:} The voice messages are heard with their labeled tones, and the receipt image is read natively.\\
\textbf{Result:} Happy tone selects the black 10x42 pair $\rightarrow$ reads \$189.99 from the receipt under that pair. \textbf{Predicted:} \textit{``\$189.99''}.
\end{tcolorbox}

\subsection{Example of Model Failure}

\textbf{Question:} \textit{``After I sounded unhappy about the wedding postponement, what price did I list the Ansel Adams print for in the photo?''}

\textbf{Expected:} \textit{``\$450''}

\textbf{Predicted (RAG MM Gemini 3.5):} \textit{``I don't know.''} \hspace{1em} \textbf{Score: 0.0}

The model failed to chain two pieces of cross-modal evidence: (1) an audio message with unhappy tone about a wedding postponement, and (2) a photo of an Ansel Adams print listing with a price. The multi-hop chain required connecting the temporal ``after'' relationship across audio prosody and image text, which the model could not complete.

\textbf{Question:} \textit{``How did my living room wall change in the video I took while repainting?''}
(\texttt{mm\_knowledge\_update\_bb832271\_001}; type \textsc{Knowledge Update})

\textbf{Expected:} \textit{``from pale yellow with peeling paint to soft sage green''}

\textbf{Predicted (ReAct MM MiMo):} \textit{``sage green''} \hspace{1em} \textbf{Score: 0.5 (Partial)}

The item is a two-session knowledge update: an earlier image session shows the
old wall colour behind a herb garden (\textsc{Selector}, background detail),
and a later video session shows the repaint (\textsc{Answer}, state-transition
gap). The model reported only the new colour, missing the required
old$\rightarrow$new contrast.

\clearpage
\section{Key Prompts}
\label{app:prompts}

This appendix documents the key LLM prompts used throughout the MOSAIC pipeline and evaluation framework.

\subsection{Evaluation Prompts}

\begin{AIbox}{RAG Answer Prompt (Text \& Multimodal)}
\textbf{System:} You are an evaluation assistant. Your task is to answer questions about a user based on retrieved conversation history [and attached media].

Answer rules:
\begin{enumerate}[leftmargin=*,itemsep=0pt,topsep=0pt]
\item Answer using ONLY the retrieved evidence provided. Do not use outside knowledge.
\item The evidence may include native multimodal artefacts (images, audio recordings, video clips). Examine each artefact carefully.
\item Your answer should be as short as possible---typically a single phrase, number, name, or date.
\item Do not include reasoning, explanations, or hedging in your final answer. Extract the requested fact directly.
\item If the retrieved evidence does not contain enough information to answer the question, respond with exactly: I don't know.
\item If the evidence contains updates or contradictions, the most recent information takes precedence.
\item For temporal references (``currently'', ``now'', ``recently''), interpret them relative to the question's date.
\end{enumerate}

\textbf{User:} \texttt{<RETRIEVED CONVERSATION HISTORY> [chunks with scores, timestamps, modalities] Question: \{question\} Answer:}
\end{AIbox}

\begin{AIbox}{ReAct System Prompt (Text Indexes --- Function Calling)}
\textbf{System:} You are a memory retrieval agent. A user asks a question about their conversation history. You have a \texttt{search\_memory} tool to find relevant information.

Rules:
\begin{itemize}[leftmargin=*,itemsep=0pt,topsep=0pt]
\item If the question can be answered without searching, answer immediately.
\item If you need more context, search with a specific, keyword-rich query.
\item Search using short queries (2--5 words), not full sentences.
\item After 2--3 searches, synthesize an answer from what you've found.
\item Answer using ONLY the retrieved evidence. Do not use outside knowledge.
\item Your answer should be as short as possible---a single phrase, number, name, or date.
\item Do not include reasoning, explanations, or hedging.
\item If the evidence contains updates or contradictions, the most recent information takes precedence.
\item Keep searching with different queries. Try different angles if earlier searches miss.
\item The question may require multiple pieces of evidence (multi-hop). Search for DIFFERENT evidence if stuck.
\end{itemize}

\textbf{Tool:} \texttt{search\_memory(query: str)} --- Search the user's conversation history for information relevant to the question. Returns ranked text excerpts.
\end{AIbox}

\begin{AIbox}{ReAct System Prompt (Native-MM --- SEARCH/ANSWER Markers)}
\textbf{System:} You are a multimodal memory retrieval agent. A user asks a question about their conversation history. You can search memory for relevant information---search results may include text, images, audio, and video files.

To search, write exactly: \texttt{SEARCH: <your query>}

To answer, write exactly: \texttt{ANSWER: <your answer>}

Rules:
\begin{itemize}[leftmargin=*,itemsep=0pt,topsep=0pt]
\item If the question can be answered immediately, write ANSWER right away. Otherwise, search first.
\item Write only ONE SEARCH or ONE ANSWER per response.
\item Search with short keyword queries (2--5 words), not full sentences.
\item Look carefully at images and listen to audio---they may contain visual or auditory details the text does not describe.
\item After 2--3 searches, synthesize an answer from what you've found.
\item Answer using ONLY the retrieved evidence. Do not use outside knowledge.
\item Your answer should be as short as possible---a single phrase, number, name, or date.
\item Do not include reasoning, explanations, or hedging in your final ANSWER.
\item Do NOT repeat the same search query.
\item Keep searching with different queries. If one search misses, try a different angle.
\end{itemize}
\end{AIbox}

\begin{AIbox}{LLM-as-Judge Prompt (5-Band Scoring Rubric)}
\label{prompt:judge}
\textbf{System:} You are an expert evaluator for a memory benchmark. Score how well a predicted answer matches the expected answer.

Score bands:
\begin{itemize}[leftmargin=*,itemsep=0pt,topsep=0pt]
\item \textbf{1.0} --- PERFECT. Semantically identical, only formatting differs. \textit{Ex: ``8:15 AM'' vs ``8:15 in the morning''}
\item \textbf{0.7} --- MOSTLY CORRECT. Right answer with extra context. \textit{Ex: ``a snowboard'' vs ``A snowboard with multicolored graphic design''}
\item \textbf{0.5} --- PARTIAL. Gets category/one part right, misses specifics. \textit{Ex: ``from \$12 to \$85'' vs ``\$85''}
\item \textbf{0.3} --- ATTEMPTED. Core fact wrong but shows effort. \textit{Ex: ``8:15 AM'' vs ``around noon''}
\item \textbf{0.0} --- WRONG. Completely incorrect, refused on answerable question, or error.
\end{itemize}

For unanswerable questions, ``I don't know'' or similar refusal scores 1.0; any other answer scores 0.0.

Respond with JSON: \texttt{\{\"reasoning\": \"...\", \"score\": <float>\}}

\textbf{User:} Question: \{q\} $|$ Answerability: \{a\} $|$ Expected: \{e\} $|$ Aliases: \{a\} $|$ Predicted: \{p\} $|$ Score this answer.
\end{AIbox}

\newpage
\subsection{Pipeline and Retrieval Prompts}

\begin{AIbox}{Query Rewrite Prompt (Keyword Extraction)}
\textbf{System:} You are a query rewriter for a conversation memory search engine. Given a question, extract the most important KEYWORDS for searching the conversation history. Return ONLY comma-separated keywords---no explanations, no full sentences.

Focus on: entity names, key attributes (colours, materials, times), actions or events mentioned, any sensory details (happy tone, photo, voice message).

Examples:
\begin{itemize}[leftmargin=*,itemsep=0pt,topsep=0pt]
\item Q: ``What was the work start time of the ship's telegraph I sounded happy about?'' $\rightarrow$ Keywords: ship telegraph, brass-and-glass, happy, work start time, clock
\item Q: ``In the photo of my road atlas, what state map is tucked in the spiral binding?'' $\rightarrow$ Keywords: road atlas photo, spiral binding, state map, tucked in
\end{itemize}

\textbf{User:} Q: \{question\} $|$ Keywords:
\end{AIbox}

\begin{AIbox}{Answer Formatting Prompt (ReAct Output Cleanup)}
\textbf{System:} You are an answer extraction assistant. Extract the final answer from the text below.

Rules:
\begin{itemize}[leftmargin=*,itemsep=0pt,topsep=0pt]
\item Your answer should be as short as possible---a single phrase, number, name, or date. Includes ``from X to Y'' for change/update answers.
\item Do not include reasoning, explanations, or hedging.
\item If the text indicates the answer is unknown, respond with: I don't know.
\item Examples: Text ``...the time was 8:15 AM...'' $\rightarrow$ ``8:15 AM'' $|$ Text ``...from \$12.00 to \$85.00...'' $\rightarrow$ ``from \$12.00 to \$85.00''
\end{itemize}

\textbf{User:} Question: \{q\} $|$ Text: \{verbose answer\} $|$ Extracted answer:
\end{AIbox}

\begin{AIbox}{LLM Quality-Judge Prompt (Four Checks)}
\textbf{System:} Evaluate this memory benchmark question for the following quality checks. Answer each with PASS or FAIL followed by a short reason.

Question type: \{qt\} $|$ Question: ``\{q\}'' $|$ Expected answer: ``\{answer\}''

Evidence facts (with modality, gap strategy, required flag, and fact statement):
\{facts\}

\begin{enumerate}[leftmargin=*,itemsep=0pt,topsep=0pt]
\item $atomicity$ Does this question ask exactly ONE thing with ONE short answer?
\item $grounding$ Is the expected\_answer factually present in the evidence facts?
\item $multimodal \_ required$ Does answering REQUIRE engaging with multimodal evidence?
\item $gap\_load\_bearing$ Do the gap strategies on required facts actually contribute to answering?
\end{enumerate}

Return JSON: \texttt{\{\"checks\": [\{\"check\": \"<id>\", \"result\": \"PASS\"|\"FAIL\", \"reason\": \"...\"\}, ...]\}}
\end{AIbox}

\end{document}